Accepted in Artificial Intelligence Review (DOI: https://doi.org/10.1007/s10462-023-10595-0)

# A survey of the Vision Transformers and their CNN-Transformer based Variants

*Asifullah Khan*[1, 2, 3*], *Zunaira Rauf*[1, 2], *Anabia Sohail*[1, 4], *Abdul Rehman Khan*[1], *Hifsa Asif*[1, 5], *Aqsa Asif*[1, 5], *and Umair Farooq*[1]

[1]Pattern Recognition Lab, Department of Computer & Information Sciences, Pakistan Institute of Engineering & Applied Sciences, Nilore, Islamabad 45650, Pakistan

[2]PIEAS Artificial Intelligence Center (PAIC), Pakistan Institute of Engineering & Applied Sciences, Nilore, Islamabad 45650, Pakistan

[3]Center for Mathematical Sciences, Pakistan Institute of Engineering & Applied Sciences, Nilore, Islamabad 45650, Pakistan

[4]Department of Electrical Engineering and Computer Science, Khalifa University of Science and Technology, Abu Dhabi, UAE

[5]Air University, E-9, Islamabad 44230, Pakistan

**Corresponding Authors:** [*]Asifullah Khan, asif@pieas.edu.pk,

## Abstract

Vision transformers have become popular as a possible substitute to convolutional neural networks (CNNs) for a variety of computer vision applications. These transformers, with their ability to focus on global relationships in images, offer large learning capacity. However, they may suffer from limited generalization as they do not tend to model local correlation in images. Recently, in vision transformers hybridization of both the convolution operation and self-attention mechanism has emerged, to exploit both the local and global image representations. These hybrid vision transformers, also referred to as CNN-Transformer architectures, have demonstrated remarkable results in vision applications. Given the rapidly growing number of hybrid vision transformers, it has become necessary to provide a taxonomy and explanation of these hybrid architectures. This survey presents a taxonomy of the recent vision transformer architectures and more specifically that of the hybrid vision transformers. Additionally, the key features of these architectures such as the attention mechanisms, positional embeddings, multi-scale processing, and convolution are also discussed. In contrast to the previous survey papers that are primarily focused on individual vision transformer architectures or CNNs, this survey uniquely emphasizes the emerging trend of hybrid vision transformers. By showcasing the potential of hybrid vision transformers to deliver exceptional performance across a range of computer vision tasks, this survey sheds light on the future directions of this rapidly evolving architecture.

# 1. Introduction

Digital images are complex in nature and exhibit high-level information, such as objects, scenes, and patterns (Khan et al. 2021a). This information can be analyzed and interpreted by computer vision algorithms to extract meaningful insights about the image content, such as recognizing objects, tracking movements, extracting features, etc. Computer vision has been an active area of research due to its applications in various fields (Bhatt et al. 2021). However, extracting high-level information from image data can be challenging due to variations in brightness, pose, background clutter, etc.

The emergence of convolutional neural networks (CNNs) has brought about a revolutionary transformation within the realm of computer vision. These networks have been successfully applied to a diverse range of computer vision tasks (Liu et al. 2018; Khan et al. 2020, 2022, 2023; Zahoor et al. 2022), especially image recognition (Sohail et al. 2021a; Zhang et al. 2023a), object detection (Rauf et al. 2023), and segmentation (Khan et al. 2021c). CNNs gained popularity due to their ability to automatically learn features and patterns from raw images (Simonyan and Zisserman 2014; Agbo-Ajala and Viriri 2021). Generally, local patterns, known as feature motifs are systematically distributed throughout the images. Different filters in the convolutional layers are specified to capture diverse feature motifs, while pooling layers in the CNNs are utilized for dimensionality reduction and to incorporate robustness against variations. This local-level processing of CNNs may result in a loss of spatial correlation, which can impact their performance when dealing with larger and more complex patterns.

Recently in computer vision, there has been some shift toward transformers, after they were first introduced by Vaswani et al. in 2017 for text processing applications (Vaswani et al. 2017a). In 2018, Parmer et al., exploited transformers for image recognition tasks, where they demonstrated outstanding results (Parmar et al. 2018). Since then, there has been a growing interest in applying transformers to various vision-related applications (Liu et al. 2021b). In 2020, Dosovitskiy et al., introduced a transformer architecture, Vision Transformer (ViT), specifically designed for image analysis, which showed competitive results (Dosovitskiy et al. 2020). ViT models work by dividing an input image into a certain number of patches, each patch is subsequently flattened and fed to a sequence of transformer layers. The transformer layers enable the model to learn the relationships between the patches and their corresponding features, allowing it to identify feature motifs on a global scale in the image. Unlike CNNs that have a local receptive field, ViTs utilize its self-attention module to model long-range relationships, which enables them to capture the global view of an image (Ye et al. 2019; Guo et al. 2021). The global receptive field of ViTs helps them retain the global relationship and thus identify complex visual patterns distributed across the image (Bi et al. 2021; Wu et al. 2023b). In this context, Maurício et al. have reported that ViTs may show promising results as compared to CNNs in various applications (Zhang et al. 2021a; Maurício et al. 2023).

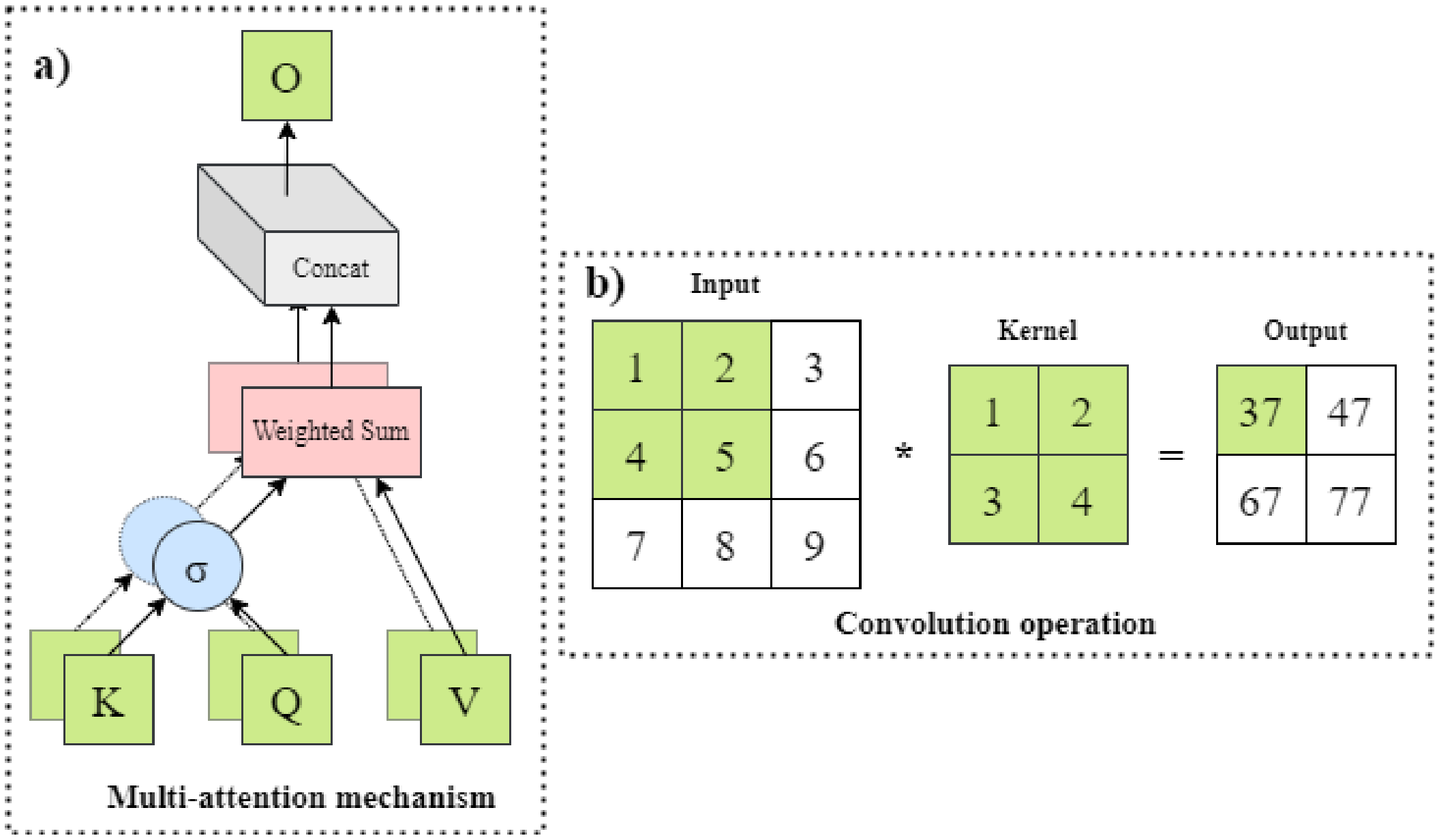

*Figure 1: Depiction of the multi self-attention (MSA) mechanism and convolution operation. MSA tends to capture global relationships, whereas the convolution operation has a local receptive filed to model pixel neighborhood information in the images.*

In addition to the difference in their design and the way of capturing visual patterns, (shown in Fig 1) CNNs and ViTs also differ in their inductive biases. CNNs heavily rely on the correlation among the neighboring pixels, whereas ViTs assume minimal prior knowledge, making them significantly dependent on large datasets (Han et al. 2023). While ViT models have produced outstanding results on object recognition, classification, semantic segmentation, and other computer vision tasks (Kirillov et al. 2023; Dehghani et al. 2023), they are not a one-size-fits-all solution. In the case of small training data, despite the large learning capacity of ViTs, they may show limited performance as compared to CNNs (Morra et al. 2020; Jamali et al. 2023). In addition, their large receptive field demands significantly more computation. Therefore, the concept of Hybrid Vision Transformers (HVT) also known as CNN-Transformer, was introduced to combine the power of both CNNs and ViTs (Maaz et al. 2023). These hybrid models leverage the convolutional layers of CNNs to capture local features, which are then fed to ViTs to gain global context using the self-attention mechanism. The HVTs have shown improved performance in many image recognition tasks.

Recently, different interesting surveys have been conducted to discuss the recent architectural and implementational advancements in transformers (Liu et al. 2021b; Du et al. 2022; Islam 2022; Aleissaee et al. 2022; Ulhaq et al. 2022; Shamshad et al. 2023). Most of these survey articles either focus on specific computer vision applications or delve into discussions on transformer models specifically developed for Natural Language Processing (NLP) applications. In contrast, this survey paper emphasizes recent developments in HVTs (CNN-Transformer), which combine concepts from both CNNs and transformers. It introduces a taxonomy for general ViTs and aims to thoroughly classify the emerging approaches based on their core architectural designs. Furthermore, this survey also conducts an in-depth analysis of HVTs, providing a taxonomy grounded in their architectural composition as well as exploring their diverse applications.

The paper begins with an introduction to the essential components of the ViT networks and then discusses various recent ViT architectures. The reported ViT models are broadly classified into six categories based on their distinct features. Additionally, a detailed discussion on HVTs is included, highlighting their focus on leveraging the advantages of both convolutional operations and multi-attention mechanisms. The survey paper covers the recent architectures and applications of HVTs in various computer vision tasks. Moreover, a taxonomy is presented for HVTs, classifying them based on the way these architectures incorporate convolution operations in combination with self-attention mechanisms. This taxonomy divides HVTs into seven major groups, each of which reflects a different way of taking advantage of both the convolutional and multi-attention operations. Frequently used abbreviations are listed in Table 1.

The paper is structured as follows: (illustrated in Fig. 2) Section 1 presents a systematic understanding of the ViT architecture, highlighting its dissimilarities with CNNs and the advent of HVT architectures. Moving on, section 2 covers the fundamental concepts used in different ViT variants, while section 3 and section 4 provide a taxonomy of the recent ViTs and HVTs architectures, respectively. Section 5 focuses on the usage of HVTs, particularly in the area of computer vision, and section 6 presents current challenges and future directions. Finally, in section 7, the survey paper is concluded.

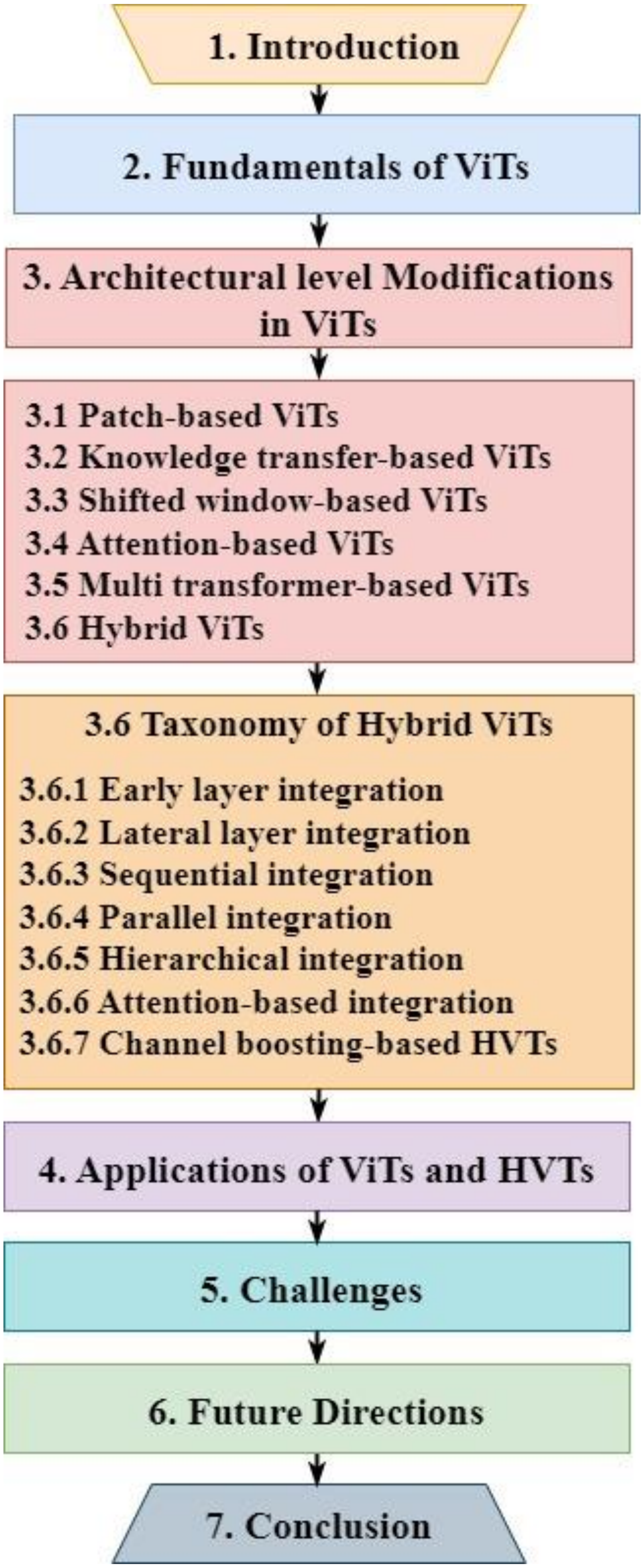

*Figure 2: Layout of the different sections of the survey paper.*

*Table 1: Table of abbreviations.*

| Abbreviation | Definition |
|---|---|
| CNN | Convolutional Neural Network |
| ViT | Vision Transformer |
| NLP | Natural Language Processing |
| HVT | Hybrid Vision Transformer |
| DL | Deep Learning |
| MSA | Multi-Head Self-Attention |
| FFN | Feed Forward Network |
| MLP | Multi-Layer Perceptron |
| APE | Absolute Position Embedding |
| RPE | Relative Position Embedding |
| CPE | Convolution Position Embedding |
| Pre-LN | Pre-Layer Normalization |
| GELU | Gaussian Error Linear Unit |
| CB | Channel Boosting |
| CvT | Convolutional Vision Transformer |
| LeFF | Locally-enhanced Feed Forward |
| CeiT | Convolution Enhanced Image Transformer |
| I2T | Image To Transformer |
| MoFFN | Mobile Feed Forward Network |
| CCT | Compact Convolutional Transformer |
| Local ViT | Local Vision Transformer |
| LeViT | LeNet-Based Vision Transformer |
| PVT | Pyramid Vision Transformer |
| MaxViT | Multi-Axis Attention-based Vision Transformer |
| MBConv | Mobile inverted bottleneck convolution |
| DPT | Deformable Patch-based Transformer |
| TNT | Transformer iN Transformer |
| DeiT | Data-efficient Image Transformer |
| TaT | Target aware Transformer |
| CaiT | Class attention in image Transformer |
| IRFFN | Inverted Residual Feed Forward Network |
| LPU | Local Perceptron Unit |
| ResNet | Residual Network |
| STE | Standard Transformer Enoder |
| SE-CNN | Squeeze and Excitation CNN |
| FPN | Feature Pyramid Network |
| UAV | Unmanned Aerial Vehicle |
| EA | Evolving Attention |
| RC | Reduction Cells |
| NC | Normal Cells |
| ConTNet | Convolution Transformer Network |
| FCT | Fully Convolutional Transformer |

## 2. Fundamental Concepts in ViTs

Fig. 3 illustrates the fundamental architectural layout of a transformer. Initially, the input image is divided, flattened, and transformed into lower dimensional linear embeddings known as Patch Embeddings. Then positional embeddings and class tokens are attached to these embeddings and fed into the encoder block of the transformer for generating class labels. In addition to the multi-head attention (MSA) layer, the encoder block contains a feed-forward neural (FFN) network, a normalization layer, and a residual connection. Finally, the last head (an MLP layer, or a decoder block) predicts the final output. Each of these components is discussed in detail in the following subsections.

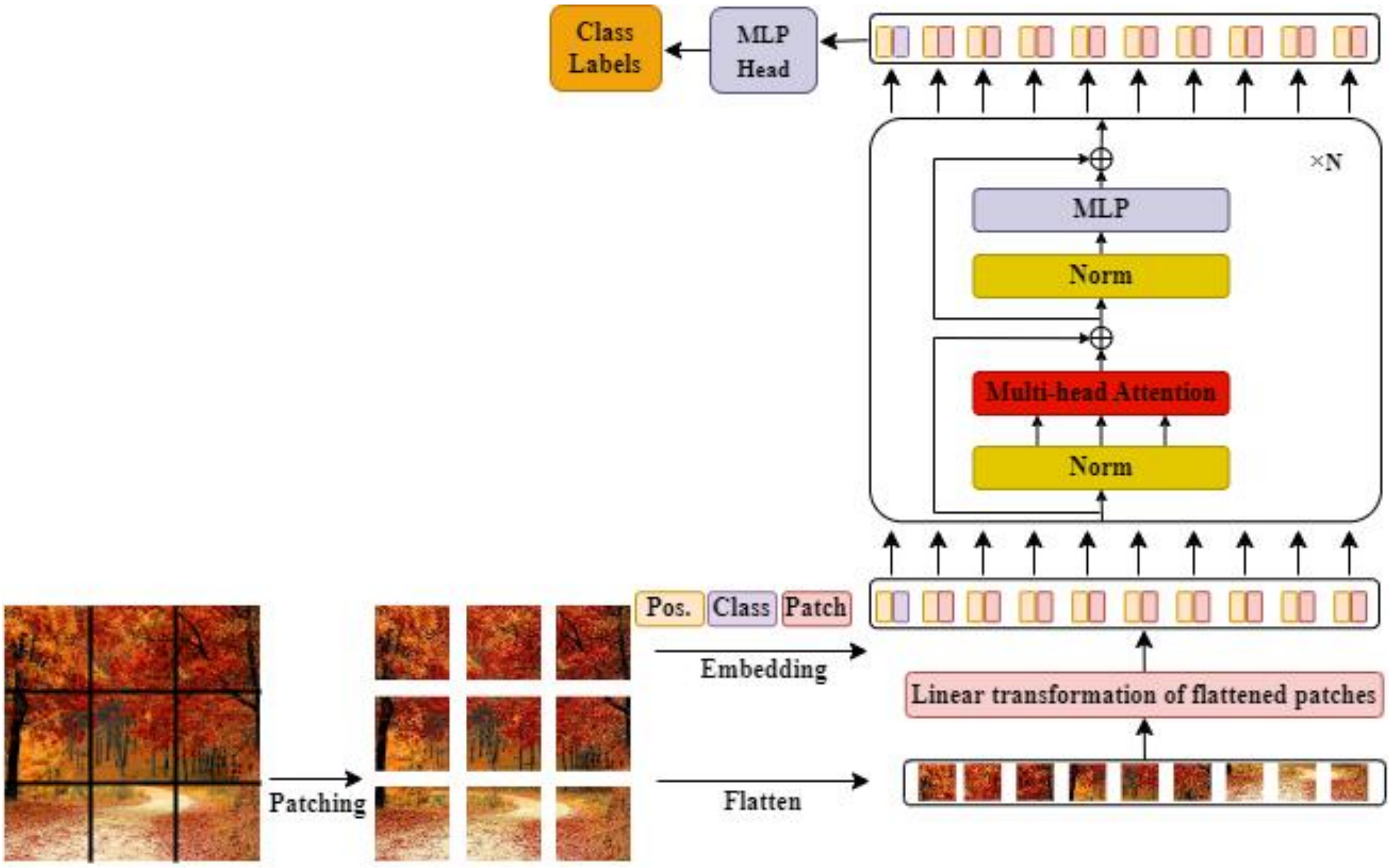

*Figure 3: Detail architecture of ViT. An input image is at first divided into patches, then their linearly transformed embeddings are combined with positional information and processed through multiple encoder/decoder blocks for downstream tasks.*

## 2.1. Patch embedding

Patch embedding is an important concept in ViT architecture. It involves converting the image patches into vector representations, which enables ViT to process images as sequences of tokens using a transformer-based approach (Dosovitskiy et al. 2020). The input image is partitioned into fixed-size non-overlapping parts, flattened into one-dimensional vectors, and projected to a higher-dimensional feature space using a linear layer with $D$ embedding dimensions (Equation 1). This approach enables ViT to learn the long-range dependencies between different patches, allowing it to attain promising results on tasks that involve images.

$$\boldsymbol{X}_{patch}^{N\times D} = R(\boldsymbol{I}_{image}^{A\times B\times C}) \quad \text{Eq. 1}$$

The input image is $\boldsymbol{I}_{image}$ with size $A \times B \times C$, $R()$ is the reshaping function to produce $N$ number of patches "$\boldsymbol{X}_{patch}$" with size $D$, and $N = \mathrm{A/P} \times \mathrm{B/P}$, $D= P \times P \times$ C, $P$ = patch size and C = channels.

## 2.2. Positional embedding

ViTs utilize positional encoding to add positional information into the input sequence and retain it throughout the network. The sequential information between patches is captured through position embeddings, which is incorporated within the patch embeddings. Since the development of ViTs, numerous position embedding techniques have been suggested for learning sequential data (Jiang et al. 2022). These techniques fall into three categories:

### 2.2.1. Absolute Position Embedding (APE)

The positional embeddings are integrated into the patch embeddings by using APE before the encoder blocks.

$$\boldsymbol{X} = \boldsymbol{X}_{patch} + \boldsymbol{X}_{pos} \quad \text{Eq. 2}$$

where, the transformer's input is represented by $X$ , $X_{patch}$ represents patch embeddings, and $X_{pos}$ is the learnable position embeddings. Both $X_{patch}$ & $X_{pos}$ have dimensions $(N+1) \times D$, where $D$ represents the dimension of an embedding. It is possible to train $X_{pos}$ corresponding to positional embeddings of a single or two sets that can be learned (Carion et al. 2020).

### 2.2.2. Relative Position Embedding (RPE)

The Relative Position Embedding (RPE) technique is primarily used to incorporate information related to relative position into the attention module (Wu et al. 2021b). This technique is based on the idea that the spatial relationships between patches carry more weight than their absolute positions. To compute the RPE value, a lookup table is used, which is based on learnable parameters. The lookup process is determined by the relative distance between patches. Although the RPE technique is extendable to sequences of varying lengths, it may increase training and testing time (Chu et al. 2021b).

### 2.2.3. Convolution Position Embedding (CPE)

The Convolutional Position Embeddings (CPE) method takes into account the 2D nature of the input sequences. 2D convolution is employed to gather position information using zero-padding to take advantage of the 2D nature (Islam et al. 2021). Convolutional Position Embeddings (CPE) can be used to incorporate positional data at different stages of the ViT. The CPE can be introduced specifically to the self-attention modules (Wu et al. 2021a), the Feed-Forward Network (FFN) (Li et al. 2021c; Wang et al. 2021b), or in between two encoder layers (Chu et al. 2021a).

## 2.3. Attention Mechanism

The core component of the ViT architecture is the self-attention mechanism, which plays a crucial role in explicitly representing the relationships between entities within a sequence. It calculates the significance of one item to others by representing each entity in terms of the global contextual information and capturing the interaction between them (Vaswani et al. 2017b). The self-attention module transforms the input sequence into three different embedding spaces namely query, key, and value. The sets of key-value pairs with query vectors are taken as inputs. The output vector is calculated by taking a weighted sum of the values followed by the softmax operator, where the weights are calculated by a scoring function (Equation 3).

$$Attention(\boldsymbol{Q}, \boldsymbol{K}, \boldsymbol{V}) = softmax\left(\frac{\boldsymbol{Q} \cdot \boldsymbol{K}^T}{\sqrt{d_k}}\right) \cdot \boldsymbol{V} \qquad \text{Eq. 3}$$

where, $\mathbf{Q}$, $\mathbf{V}$, and $\mathbf{K}^{\mathbf{T}}$ are query, value, and transposed key matrix, respectively. $\frac{1}{\sqrt{\mathrm{d_k}}}$ is the scaling factor, and $\mathrm{d_k}$ is dimensions of the key matrix.

### 2.3.1. Multi-Head Self-Attention (MSA)

The limited capacity of a single-head self-attention module often leads to its focus on only a few positions, potentially overlooking other important positions. To address this limitation, MSA is employed. MSA utilizes parallel stacking of self-attention blocks to increase the effectiveness of the self-attention layer (Vaswani et al. 2017b). It captures a diverse range of complex interactions among the sequence elements by assigning various representation subspaces (query, key, and value) to the attention layers. The MSA constitutes multiple self-attention blocks. Each is equipped with learnable weight matrices for query, key, and value sub-spaces (Equation 4). The outputs of these blocks are then concatenated and projected to the output space using the learnable parameter $W^O$ (Equation 5)This enables the MSA to focus on multiple portions and to effectively capture the relationships in all areas. The mathematical representation of the attention process is given below:

$$head_i = Attention(\boldsymbol{Q}_i, \boldsymbol{K}_i, \boldsymbol{V}_i),\ and\ i = 1,2, \dots, h \quad \text{Eq. 4}$$

$$MSA(\boldsymbol{Q}, \boldsymbol{K}, \boldsymbol{V}) = Concat(head_1, head_2, \cdots, head_h) \cdot \boldsymbol{W}^O \quad \text{Eq. 5}$$

In Eq. 4, $\text{head}_\text{i}$ is the output of each self-attention block, whereas $\mathbf{W}^\text{O}$ in Eq.5 represents the learnable parameters.

Self-attention's capability to dynamically compute filters for every input sequence is a significant advantage over convolutional processes. Unlike convolutional filters, which are often static, self-attention can adjust to the particular context of the input data. Self-attention is also robust to changes in the number of input points or their permutations, which makes it a good choice for handling irregular inputs. Traditional convolutional procedures, on the other hand, are less adaptable to handling inputs with variable objects and require a grid-like structure, like 2D images. Self-attention is a powerful tool for modeling sequential data and has been effective in various tasks including NLP (Khan et al. 2021b).

## 2.4. Transformer layers

A ViT encoder consists of several layers to process the input sequence. These layers comprise the MSA mechanism, feed-forward neural network (FFN), residual connection, and normalization layer. These layers are arranged to create a unified block that is repeated several times to learn the complex representation of the input sequence.

### 2.4.1. Feed-forward network

A transformer-specific feed-forward network (FFN) is employed in models to obtain more complex attributes from the input data. It contains multiple fully connected layers and a nonlinear activation function, such as GELU in between the layers (Equation 6). FFN is utilized in every encoder block after the self-attention module. The hidden layer of the FFN usually has a dimensionality of 2048. These FFNs or MLP layers are local and translationally equivalent to global self-attention layers (Dosovitskiy et al. 2020).

$$FFN(\boldsymbol{X}) = b2 + \boldsymbol{W}2 * \sigma(b1 + \boldsymbol{W}1 * \boldsymbol{X}) \qquad \text{Eq. 6}$$

In Eq. 6, the non-linear activation function GELU is represented by σ. Weights of the network are represented as **W**1, and **W2**, whereas b1, and b2 correspond to layer-specific bias

### 2.4.2. Residual connection

Sub-layers in the encoder/decoder block (MSA and FFN) utilize a residual link to improve performance and strengthen the information flow. Original input positional embedding is added to the output vector of MSA, as additional information. The residual connection is then followed by a layer-normalization operation (Equation 7).

$$\boldsymbol{X}_{output} = LayerNorm(\boldsymbol{X} \oplus \boldsymbol{O}_{SL}) \qquad \text{Eq. 7}$$

Where $\boldsymbol{X}$ is the original input and $\boldsymbol{O}_{SL}$ is the output of each sub-layer, ⊕ representing the residual connection.

### 2.4.3. Normalization layer

There are various methods for layer normalization, such as pre-layer normalization (Pre-LN) (Kim et al. 2023), which is utilized frequently. The normalization layer is placed prior to the MSA or FFN and inside the residual connection. Other normalization procedures, including batch normalization, have been suggested to enhance the training of transformer models, however, they might not be as efficient due to changes in the feature values (Jiang et al. 2022).

### 2.5. Hybrid Vision Transformers (CNN-Transformer Architectures)

In the realm of computer vision tasks, ViTs have gained popularity, but compared to CNNs, they still lack image-specific inductive bias often referred to as prior knowledge (Seydi and Sadegh 2023). This inductive bias includes characteristics like translation and scale invariance due to the shared weights across different spatial locations (Moutik et al. 2023). In CNNs, the locality, translational equivariance, and two-dimensional neighborhood structure are ingrained in every layer throughout the whole model. Additionally, the kernel leverages the correlation between neighboring pixels, which facilitates the extraction of good features quickly (Woo et al. 2023). On the other hand, in ViT, the image is split into linear patches (tokens) that are fed into encoder blocks through linear layers to model global relationships in the images. However, linear layers lack effectiveness in extracting local correlation (Woo et al. 2023).

Many HVT designs have focused on the efficiency of convolutions in capturing local features in images, especially at the start of the image processing workflow for patching and tokenization (Guo et al. 2023). The Convolutional Vision Transformer (CvT), for instance, uses a convolutional projection to learn the spatial and low-level information in image patches. It also utilizes a hierarchical layout with a progressive decrease in token numbers and an increase in token width to mimic the spatial downsampling effect in CNNs (Wu et al. 2021a). Similarly, Convolution-enhanced Image Transformers (CeiT) leverage convolutional operations to extract low-level features via an image-to-token module (Yuan et al. 2021a). A novel sequence pooling technique is presented by the Compact Convolutional Transformer (CCT), which also integrates conv-pool-reshape blocks to carry out tokenization (Hassani et al. 2021). It also showed an accuracy of about 95% on smaller datasets like CIFAR10 when trained from scratch, which is generally difficult for other traditional ViTs to achieve.

Several recent studies have investigated ways to enhance the local feature modeling capabilities of ViTs. LocalViT employs depthwise convolutions to improve the ability to model local features (Li et al. 2021c). LeViT uses a CNN block with four layers at the beginning of the ViT architecture to gradually increase channels and improve efficiency at inference time (Graham et al. 2021). Similar methods are employed by ResT, however, to manage fluctuating image sizes, depth-wise convolutions, and adaptive position encoding are used (Zhang and Yang 2021).

Without additional data, CoAtNets' unique architecture of depthwise convolutions and relative self-attention achieves outstanding ImageNet top-1 accuracy (Dai et al. 2021). In order to create stronger cross-patch connections, Shuffle Transformer provides a shuffle operation (Huang et al. 2021b) and CoaT is a hybrid approach that incorporates depthwise convolutions and cross-attention to encode relationships between tokens at various scales (Xu et al. 2021a). Another method "Twins" builds upon PVT by incorporating separable depthwise convolutions and relative conditional position embedding (Chu et al. 2021a). Recently, MaxVit, hybrid architecture, introduced the idea of multi-axis attention. Their hybrid block consists of MBConv-based convolution followed by block-wise self-attention and grid-wise self-attention, and when repeated multiple times this block creates a hierarchical representation and is capable of tasks like image generation and segmentation (Tu et al. 2022b). The block-wise and grid-wise attention layers are capable of extracting local and global features respectively. Convolution and transformer model strengths are intended to be combined in these hybrid designs.

# 3. Architectural level modifications in ViTs

In recent years, different modifications have been carried out in ViT architectures (Zhou et al. 2021). These modifications can be categorized based on their attention mechanism, positional encoding, pre-training strategies, architectural changes, scalability, etc. ViT architectures can be broadly classified into five main classes based on the type of architectural modification, namely, (i) patch-based approaches, (ii) knowledge transfer-based approaches, (iii) shifted window-based approaches, (iv) attention-based approaches, and (v) multi-transformer-based approaches. However, it is observed that with the introduction of CNN's inductive bias to ViTs there came a boost in its performance. In this regard, we also classified the HVTs into seven categories based on their structural design. The taxonomy of ViT architectures is shown in Fig. 4. In addition a comprehensive overview of various online resources relevant to ViTs including libraries, lecture series, datasets, and computing platforms are provided in Table 2.

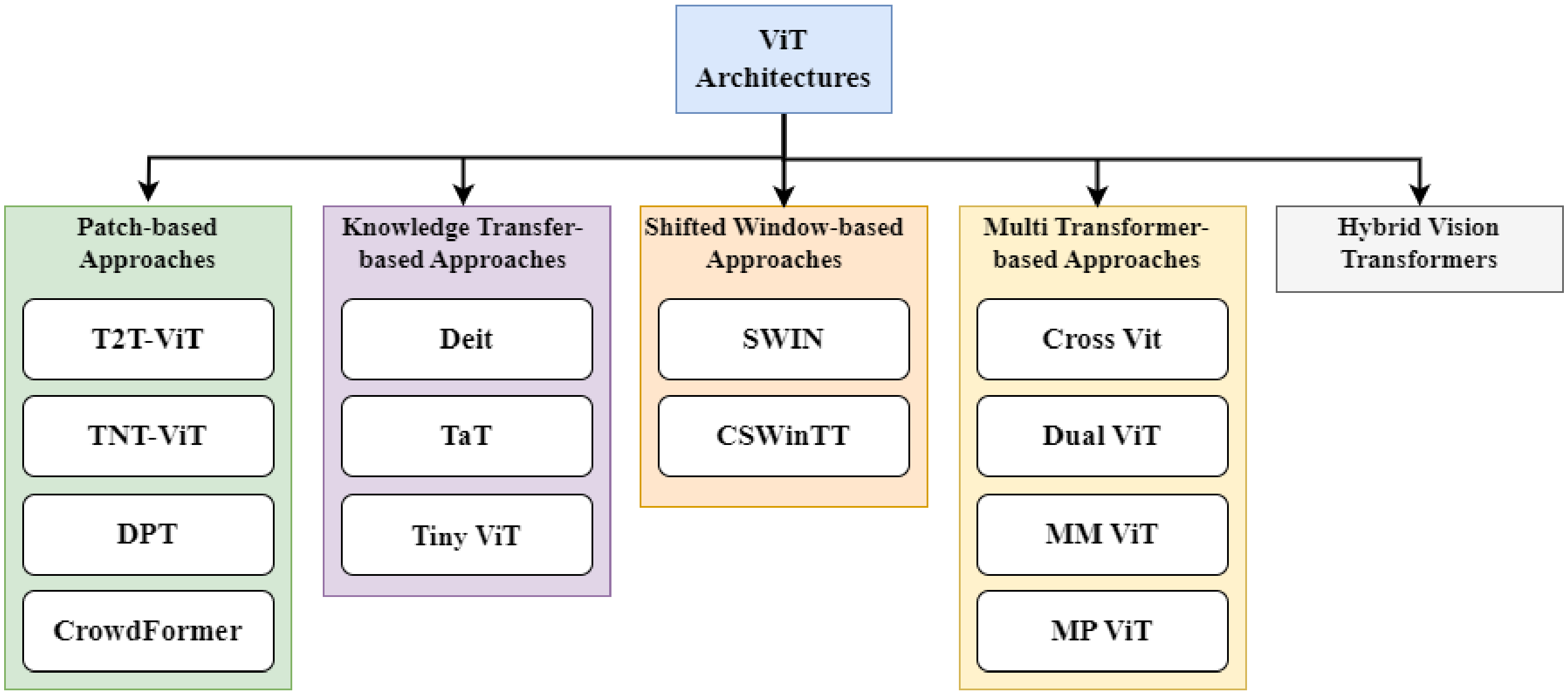

*Figure 4: Taxonomy of Vision ViT architectures.*

*Table 2: Distinct online resources relevant to DL and ViT.*

<table>
<tr><th>Category</th><th>Description</th><th>Source</th></tr>
<tr><td rowspan="7">Cloud Computing Solutions</td><td rowspan="2">Online available datasets and computational resources</td><td>Google Colab:<br>https://colab.research.google.com/notebooks/welcome.ipynb</td></tr>
<tr><td>Kaggle:<br>https://www.kaggle.com/</td></tr>
<tr><td rowspan="5">DL commercially available services</td><td>FloydHub:<br>https://www.floydhub.com/</td></tr>
<tr><td>Amazon SageMaker:<br>https://aws.amazon.com/deep-learning/</td></tr>
<tr><td>Microsoft Azure ML Services:<br>https://azure.microsoft.com/en-gb/services/machine- learning/</td></tr>
<tr><td>Google Cloud:<br>https://cloud.google.com/deep-learning-vm/</td></tr>
<tr><td>IBM Watson Studio:<br>https://www.ibm.com/cloud/deep-learning</td></tr>
<tr><td rowspan="8">DL Libraries</td><td rowspan="8">DL libraries with integrated neural network classes and optimization for CPU and GPU</td><td>Pytorch:<br>https://pytorch.org/</td></tr>
<tr><td>Tensorflow:<br>https://www.tensorflow.org/</td></tr>
<tr><td>MatConvNet:<br>http://www.vlfeat.org/matconvnet/</td></tr>
<tr><td>Keras:<br>https://keras.io/</td></tr>
<tr><td>Theano:<br>http://deeplearning.net/software/theano/</td></tr>
<tr><td>Caffe:<br>https://caffe.berkeleyvision.org/</td></tr>
<tr><td>Detectron2:<br>https://github.com/facebookresearch/detectron2</td></tr>
<tr><td>OpenMMLAB:<br>https://openmmlab.com/</td></tr>
<tr><td rowspan="4">Lecture Series</td><td rowspan="4">Free online courses on DL</td><td>Stanford Lecture Series:<br>https://web.stanford.edu/class/cs224n/</td></tr>
<tr><td>Youtube:<br>https://www.youtube.com/watch?v=SZorAJ4I-sA</td></tr>
<tr><td>Youtube:<br>https://www.youtube.com/watch?v=iDulhoQ2pro</td></tr>
<tr><td>Coursera:<br>https://www.coursera.org/learn/nlp-sequence-models</td></tr>
<tr><td rowspan="2">Datasets</td><td rowspan="2">Diverse categories of annotated image datasets available online for free access</td><td>ImageNet:<br>http://image-net.org/</td></tr>
<tr><td>COCO:<br>http://cocodataset.org/#home</td></tr>
</table>

| | | |
|---|---|---|
| | | **Visual Genome:** http://visualgenome.org/ |
| | | **Open images:** https://ai.googleblog.com/2016/09/introducing-open-images-dataset.html |
| | | **Places:** https//places.csail.mit.edu/index.html |
| | | **Youtube-8M:** https://research.google.com/youtube8m/index.html |
| | | **CelebA:** http://mmlab.ie.cuhk.edu.hk/projects/CelebA.html |
| | | **Wiki Links:** https://code.google.com/archive/p/wiki-links/downloads |
| | | **EXCITEMENT dataset:** https://github.com/hltfbk/EOP-1.2.1/wiki/Data-Sets |
| | | **Ubuntu Dialogue Corpus:** https://www.kaggle.com/datasets/rtatman/ubuntu-dialogue-corpus |
| | | **ConvAI3:** https://github.com/DeepPavlovAdmin/convai |
| | | **Large Movie Review Dataset:** https://ai.stanford.edu/~amaas/data/sentiment/ |
| | | **CIFAR10:** https://www.cs.toronto.edu/~kriz/cifar.html |
| | | **Indoor Scene Recognition:** http://web.mit.edu/torralba/www/indoor.html |
| | | **Computer Vision Datasets:** https://computervisiononline.com/datasets |
| | | **MonuSeg:** https://monuseg.grand-challenge.org/Data/ |
| | | **Oxford-IIIT Pets:** https://www.robots.ox.ac.uk/~vgg/data/pets/ |
| | | **Fashion MNIST:** https://research.zalando.com/welcome/mission/research-projects/fashion-mnist |
| **Blogs/ Repositories** | **High-quality, free online articles and blogs** | **Github.io:** http://jalammar.github.io/illustrated-transformer/ |
| | | **Github:** https://github.com/huggingface/pytorch-image-models |
| | | **Viso Ai:** https://viso.ai/deep-learning/vision-transformer-vit/ |
| | | **Github:** https://github.com/google-research/vision_transformer |
| | | **HuggingFace:** https://huggingface.co/docs/transformers/model_doc/vit |
| **Hardware Resources** | **Energy-efficient and computationally optimized hardware solutions for DL processing** | **NVIDIA:** http://nvdla.org/ |
| | | **FPGA:** https://www.intel.com/content/www/us/en/artificial-intelligence/programmable/overview.html |
| | | **Eyeriss:** http://everiss.mit.edu/ |
| | | **AMD:** https://www.amd.com/en/graphics/instinct-server-accelerators |
| | | **Google's TPU:** https://cloud.google.com/tpu/ |

## 3.1. Patch-based approaches

In ViT, an image is first divided into a grid of patches, which are subsequently flattened to generate linear embedding, treated as a sequence of tokens (Dosovitskiy et al. 2020). Positional embedding and class tokens are added to these embeddings, which are then given to the encoder for feature learning. Several studies exploited different ways of patch extraction mechanisms to improve the performance of ViTs. These mechanisms include fixed-size patching (Wang et al. 2021c), dynamic patching (Ren et al. 2022; Zhang et al. 2022c), and overlapping patching (Wang et al. 2021b). In this regard, we discuss several architectures and their patching criteria.

### 3.1.1. Tokens-to-Token Vision Transformer (T2T-ViT)

Tokens-to-Token Vision Transformer (T2T-ViT) utilizes a fixed size and iterative approach to generate patches (Yuan et al. 2021b). It utilizes the proposed Token-to-Token module iteratively to generate patches from the images. The generated patches are then fed to the T2T-ViT network to obtain final predictions.

#### 3.1.2. Transformer in Transformer (TNT-ViT)

Transformer in Transformer ViT (TNT-ViT) presented a multi-level patching mechanism to learn representations from objects with different sizes and locations (Han et al. 2021). It first divides the input image into patches then each patch is further divided into sub-patches. Later, the architecture utilizes different transformer blocks to model the relationship between the patches and sub-patches. Extensive experiments showed the efficiency of TNT-ViT in terms of image classification on the ImageNet dataset.

#### 3.1.3. Deformable Patch-based Transformer (DPT)

Deformable Patch-based Transformer (DPT) presented an adaptive patch embedding module named DePatch (Chen et al. 2021e). Fixed-size patching in transformers results in a loss of semantic information, which affects the system's performance. In this regard, the proposed DePatch module in DPT splits the images in an adaptive way to obtain patches with variable sizes and strong semantic information.

#### 3.1.4. CrowdFormer

Yang and co-authors developed a ViT architecture, CrowdFormer for crowd counting (Yang et al. 2022b). The proposed architecture utilizes its overlap patching transformer block to capture the crowd's global contextual information. To consider images at different scales and in a top-down manner, the overlap patching layer is exploited, where instead of fixed-sized patches, a sliding window is used to extract overlapping patches. These overlapping patches tend to retain the relative contextual information for effective crowd counting.

## 3.2. Knowledge transfer-based approaches

This category enlists those ViT architectures that utilize a knowledge transfer (knowledge distillation) approach. It involves conveying knowledge from a larger network to a smaller network, much like a teacher imparting knowledge to a student (Kanwal et al. 2023; Habib et al. 2023). The teacher model is usually a complex model with ample learning capability, while the student model is simpler. The basic idea behind knowledge distillation is to facilitate the student model in acquiring and incorporating the distinctive features of the teacher model. This can be particularly useful for tasks where computational resources are limited, as the smaller ViT model can be deployed more efficiently than the larger one.

### 3.2.1. Data-efficient Image Transformers (DeiT)

Deit is a smaller and more efficient version of ViT, which has shown competitive performance on various tasks (Touvron et al. 2020). It uses a pre-trained ViT model for the teacher and a smaller version for the student. Usually, supervised and unsupervised learning is used in combination, with the teacher network supervising the student network to produce similar results. In addition to the fast inference time and limited computational resources of DeiT, it also has an improved generalization performance because the student model has learned to capture the most important features and patterns in the data, rather than just memorizing the training data.

### 3.2.2. Target-aware Transformer (TaT)

Target-aware Transformer (TaT) (Lin et al. 2022) utilized one-to-many relation to exchange information from the teacher to the student network. The feature maps were first divided into a number of patches, then for each patch, all the teacher’s features were transferred to all the student features rather than employing correlation between all spatial regions. All the features inside a patch were then averaged into a single vector to make the knowledge transfer computationally efficient.

### 3.2.3. Tiny Vision Transformer (TinyViT)

Wu et al. suggested a fast distillation methodology along with a novel architecture, known as TinyViT (Wu et al. 2022a). Their main concept was to convey the learned features of the large pre-trained models to the tiny ones during pre-training (Fig. 5). The output logits of the instructor models were reduced and stored in addition to the encoded data augmentations on the disc beforehand to save memory and computational resource. The student model then employs a decoder to re-construct the saved data augmentations and knowledge is transferred via the output logits with both the models trained independently. Results demonstrated TinyViT's effectiveness on large-scale test sets.

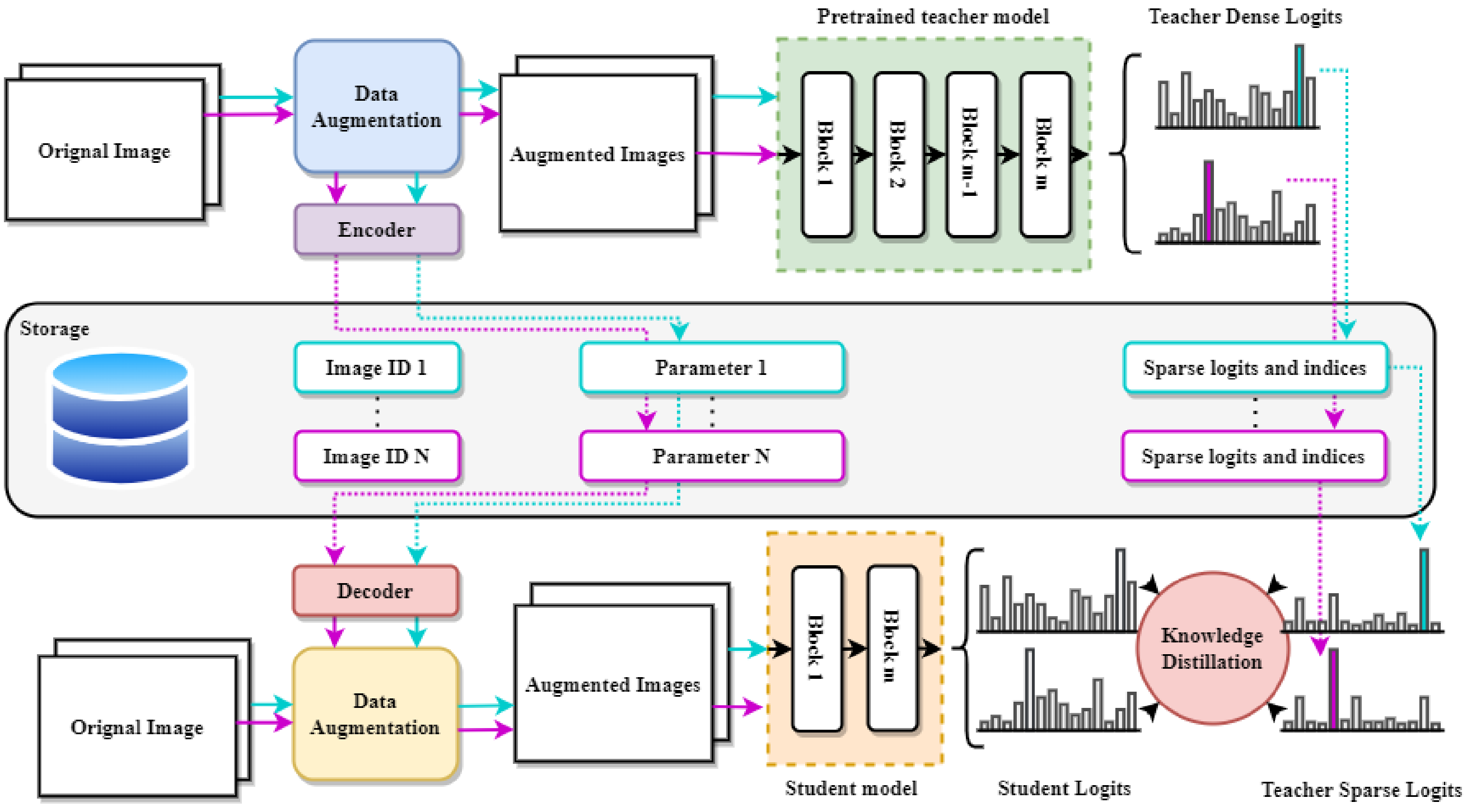

*Figure 5: Detailed workflow of knowledge transfer-based approach (TinyViT).*

## 3.3. Shifted window-based approaches

Several ViT architectures have adopted the shifted window-based approach to enhance their performance. This approach was first introduced by Liu et al. in their Swin Transformer (Liu et al. 2021c). The Swin Transformer has a similar architecture to ViT but with a shifted windowing scheme, as shown in Fig. 6. It controls the self-attention computation by computing it within each non-overlapping local window, while still providing cross-window connections to improve the efficacy. This is achieved by implementing shifted window-based self-attention as two consecutive Swin Transformer blocks. The first block uses regular window-based self-attention, and the second block shifts those windows and applies regular window-based self-attention again. The idea behind shifting the windows is to enable cross-window connections, which can help the network in improving its capability to model global relationships.

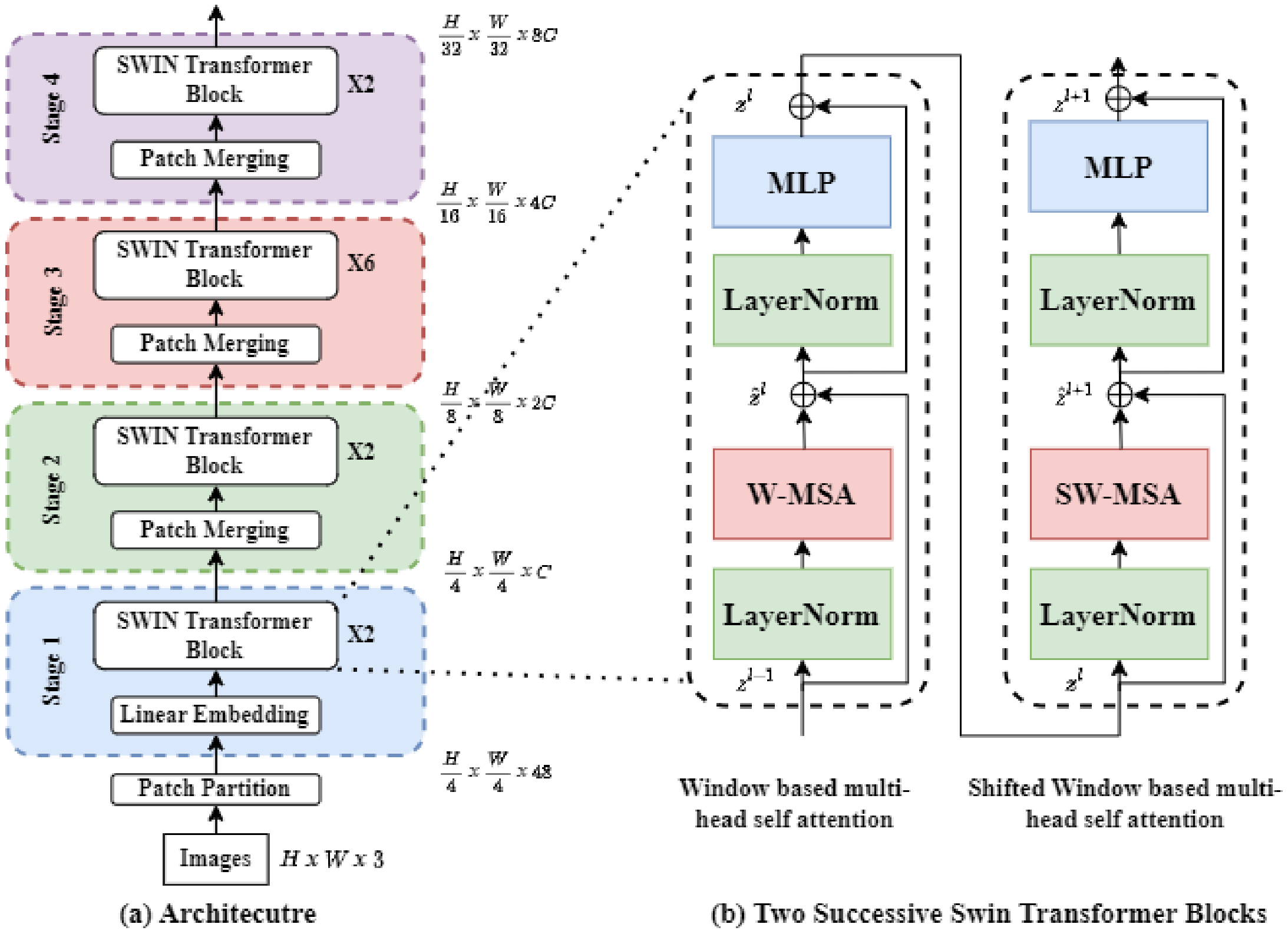

*Figure 6: Architectural diagram of Swin Transformer (shifted window-based approach).*

Song et al. proposed a novel ViT architecture for visual object tracking, named CSWinTT which utilizes cyclic shifting window-based attention at multi-scales (Song et al. 2022b). This approach enhances pixel attention to window attention and enables cross-window multi-scale attention to aggregate attention at different scales. This ensures the integrity of the tracking object and generates the best fine-scale match for the target object. Moreover, the cyclic shifting technique expands the window samples with positional information, which leads to greater accuracy and computational efficiency. By incorporating positional information into the attention mechanism, the model is better equipped to handle changes in the object's position over time and can track the object more effectively. Overall, the proposed architecture has shown promising results in improving the accuracy and efficiency of visual object tracking using ViT-based models.

## 3.4. Attention-based approaches

Numerous ViT architectures have been proposed that modify the self-attention module to enhance their performance. Some of these models utilize dense global attention mechanisms (Vaswani et al. 2017a; Dosovitskiy et al. 2020), while others utilize sparse attention mechanisms (Jiang et al. 2021; Liu et al. 2021c; Dai et al. 2021) to capture global-level dependencies in the images with no spatial correlation. These types of attention mechanisms are known to be computationally expensive. A number of works have been done to improve the attention modules in terms of performance and computational complexity (Tu et al. 2022b).

### 3.4.1. Class attention layer (CaiT)

Touvron et al. introduced a new approach to improve the performance of deep transformers (Touvron et al. 2021). Their architecture, named, CaiT contains a self-attention module and a class attention module. The self-attention module is just like a normal ViT architecture, but the class token (class information) is not added in the initial layers. The class embeddings are added in the class attention module, later in the architecture. Their approach showed good results with a few numbers of parameters.

### 3.4.2. Deformable attention transformer (DAT)

Xia and co-authors proposed a data-dependent attention mechanism to focus on the more reliable regions (Xia et al. 2022). Their architecture has a modular design with each stage having a local attention layer followed by a deformable attention layer in each stage. The proposed DAT architecture showed exemplary performance on benchmark datasets.

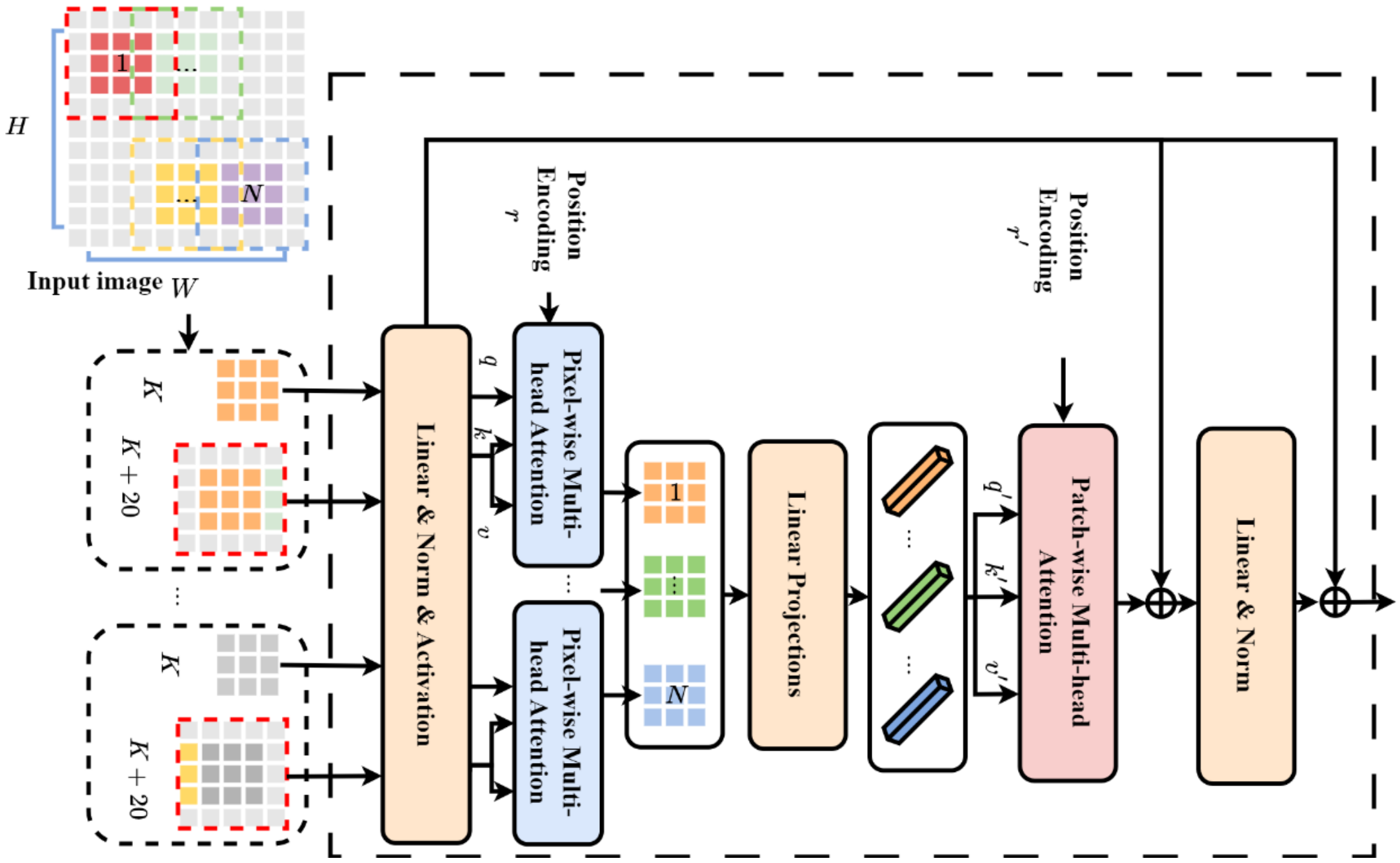

*Figure 7: Architecture of patch-based Separable Transformer (SeT), which modified its MSA layer by introducing two diverse attention blocks.*

### 3.4.3. Patch-based Separable Transformer (SeT)

Sun et al. used two different attention modules in their ViT architecture to fully capture global relationships in images (Sun et al. 2022) (Fig. 7). They proposed a pixel-wise attention module to learn local interactions in the initial layers. Later they utilized a patch-wise attention module to extract global-level information. SeT showed superior results than other methods on several datasets, including ImageNet and MS COCO datasets.

## 3.5. Multi-transformer-based approaches

Many approaches utilized multiple ViTs in their architecture to improve their performance on various tasks that require multi-scale features. This section discusses such multi-transformer-based ViT architectures.

### 3.5.1. Cross Vision Transformer (CrossViT)

Chen and co-authors proposed a ViT architecture having dual branches which they named as CrossViT (Chen et al. 2021a). The key innovation in the proposed model is the combination of image patches of different sizes, which enables CrossViT to generate highly domain-relevant features. The smaller and larger patch tokens are processed using two separate branches with varying computational complexities. The two branches are fused multiple times using an efficient cross-attention module. This module enables knowledge transfer between the branches by creating a non-patch token. The attention map generation is achieved linearly, rather than quadratically, through this process. This makes CrossViT more computationally efficient than other models that use quadratic attention.

### 3.5.2. Dual Vision Transformer (Dual-ViT)

The Dual Vision Transformer (Dual-ViT) is a new ViT architecture that reduces the computational cost of self-attention mechanisms (Yao et al.). This architecture utilizes two individual pathways to capture global and local-level information. The semantic branch learns the coarse details, whereas the pixel pathway captures more fine details in the images. both of these branches are integrated and trained in parallel. The proposed dualViT showed good results on the ImageNet dataset with fewer parameters as compared to other existing models.

### 3.5.3. Multiscale Multiview Vision Transformer (MMViT)

Multiscale Multiview Vision Transformer (MMViT) incorporates multiscale feature maps and multiview encodings into transformer models. The MMViT model utilizes several feature extraction stages to process multiple views of the input at various resolutions in parallel. At each scale stage, a cross-attention block is exploited to merge data across various perspectives. This approach enables the MMViT model to obtain high-dimensional representations of the input at multiple resolutions, leading to complex and robust feature representations.

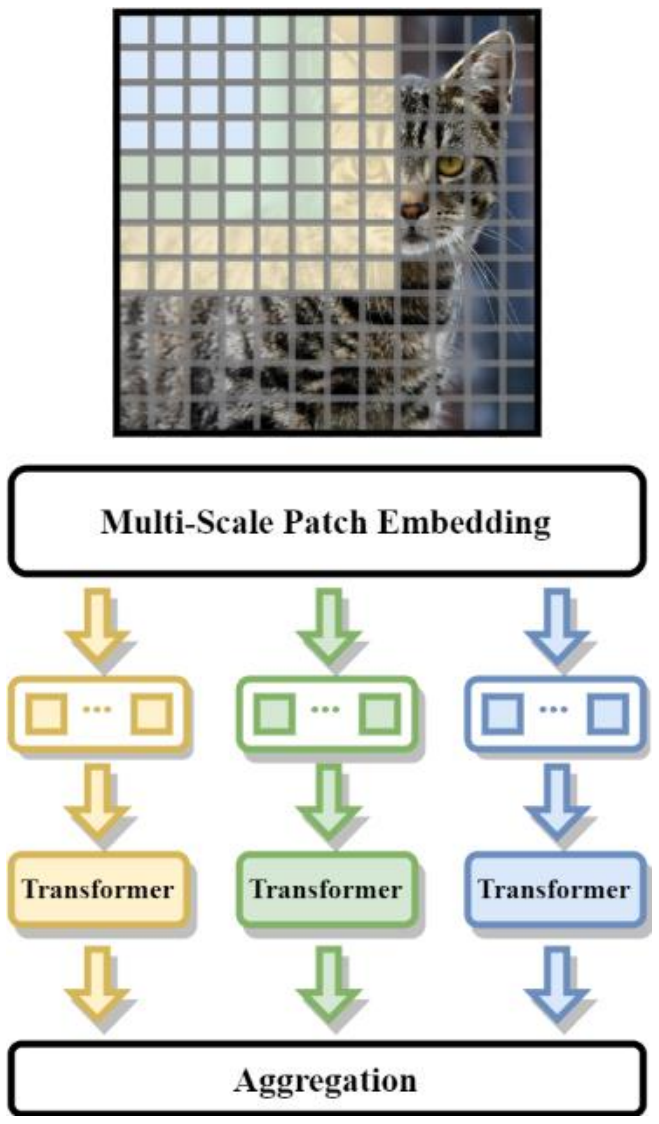

*Figure 8: Architecture of Multi-transformer-based MPViT, which utilize multiple transformers in its architecture.*

### 3.5.4. Multi-Path Vision Transformer (MPViT)

MPViT utilizes the multi-scale patching technique and multi-path-based ViT architecture to learn feature representations at different scales (Lee et al. 2021b). Their proposed multi-scale patching technique utilizes CNNs to create feature maps at different scales (Fig. 8). Later they utilize multiple transformer encoders to process multi-scale patch embeddings. Lastly, they aggregate the outputs from each encoder to generate an aggregated output. The proposed MPViT demonstrated superior results as compared to existing approaches on the ImageNet dataset.

## 3.6. Taxonomy of HVTs (CNN-Transformer architectures)

Despite their successful performance, ViTs face three main issues, a) inability to capture low-level features by considering correlation in the local neighborhood, b) expensive in terms of computation and memory consumption due to their MSA mechanism, c) and fixed-sized input tokens, embedding. To overcome these issues, there comes the boom of hybridization of CNNs and ViTs after 2021. HVTs combine the strengths of both CNNs and transformer architectures to create models for capturing both the local patterns and global context in images (Yuan et al. 2023b). They have gained valuable focus in the research community due to their promising results in several image-related tasks (Li et al. 2022). Researchers have proposed a variety of architectures in this field by exploiting different approaches to merge CNNs and transformers (Heo et al. 2021; Si et al. 2022). These approaches include but are not limited to, adding some CNN layers within transformer blocks (Liu et al. 2021a; He et al. 2023; Wei et al. 2023), introducing a multi-attention mechanism in CNNs (Zhang et al. 2021b; Ma et al. 2023b), or using CNNs to extract local features and transformers to capture long-range dependencies (Yuan et al. 2021a, 2023a; Zhang et al. 2023c). In this regard, we define some subcategories based on the pattern of integration of the convolution operation within ViT architectures. These include (1) early-layer integration, (2) lateral-layer integration, (3) sequential integration, (4) parallel integration, (5) block integration, (6) hierarchical integration, (7) attention-based integration, and (8) channel boosting integration, depicted in Fig. 9.

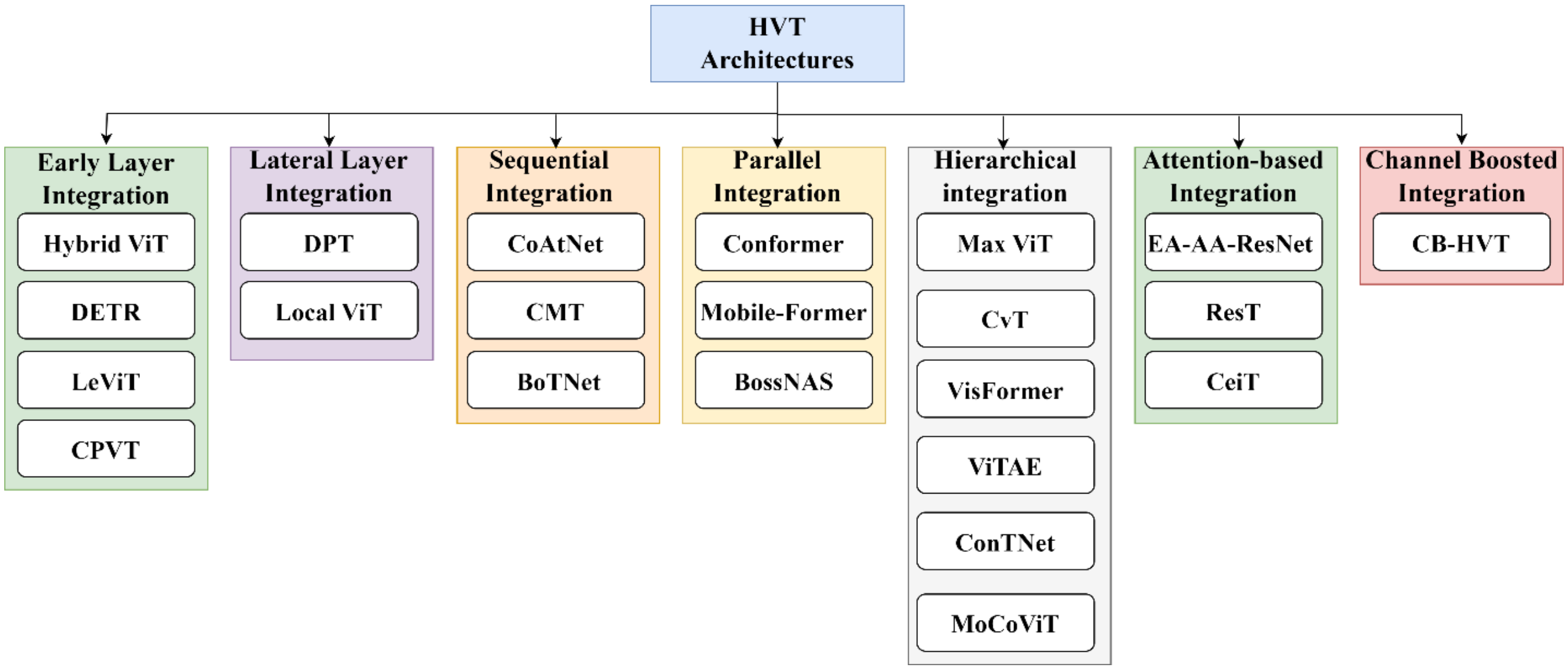

*Figure 9: Taxonomy of Hybrid ViTs.*

### 3.6.1. Early-layer integration

Long-range dependencies in the images are well-captured by ViTs, but since there is no inductive bias, training them needs a lot of data. On the other hand, CNNs inherent image-related inductive bias and capture high-level correlation present in the images locally. Therefore, researchers are focusing on designing HVTs, to merge the benefits of both CNNs and transformers (Pan et al. 2022). A lot of work is done to find out the most optimal way to fuse the convolution and attention in the transformer architectures. CNNs can be utilized at different levels to incorporate the locality in the architectures. Various studies have suggested the idea that it is beneficial to first capture local patterns and then learn the long-range dependencies to have a more optimized local and global perspective of an image (Peng et al. 2023).

***Hybrid ViT***

The first ViT architecture was proposed by Dosovitskiy *et al.* in 2020 (Dosovitskiy et al. 2020). In their work, they suggested the idea of considering image patches as sequences of tokens and feeding them into a transformer-based network to perform image recognition tasks. In their paper, they laid the foundation for HVTs by presenting a hybrid version of ViT. In the hybrid architecture, the input sequences were obtained from CNN feature maps instead of raw image patches (LeCun et al. 1989). The input sequence was created by flattening the feature maps spatially, and the patches were produced using a 1x1 filter. They utilized ResNet50 architecture to obtain the feature maps as input to ViT (Wu et al. 2019). In addition, they carried out extensive experiments to identify the optimal intermediate block for feature map extraction.

### *Detection Transformer (DETR)*

Carion et al. proposed a Detection Transformer (DETR) for performing object detection in natural images in 2020 (Carion et al. 2020). In their end-to-end proposed approach, they initially utilized a CNN to process the input before feeding it to the ViT architecture. The feature maps from the CNN backbone were combined with fixed-sized positional embeddings to create input for the ViT encoder. The outputs from the ViT decoder were then fed to a feed-forward network to make final predictions. DETR showed better performance when compared to other revolutionary detection models like Faster R-CNN. Their detailed idea is depicted in Fig. 10.

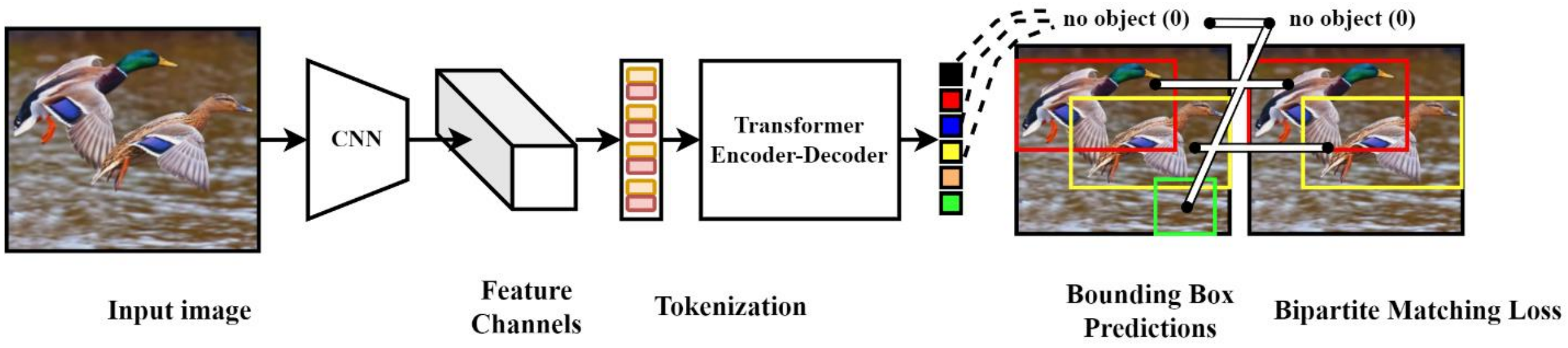

*Figure 10: Architecture of DETR, with CNN integration as an initial stem block.*

### *LeNet-based Vision Transformer (LeViT)*

Graham et al. proposed a hybrid ViT "LeViT" in 2021 (Graham et al. 2021). In their model, they utilized convolution layers initially for processing the input. The proposed architecture combined a CNN with the MSA of ViT architecture to extract local as well as global features from the input image. The LeViT architecture at first utilized a four-layered CNN model to reduce the image resolution and to obtain local feature representations. These representations were then fed to a ViT-inspired multi-stage architecture with MLP and attention layers to generate output.

### *Conditional Positional Encodings for Vision Transformers (CPVT)*

CPVT was proposed by Chu et al. in 2023 (Chu et al. 2021b). In their work, they devised a new scheme of conditional positional embeddings to improve the performance of ViTs (Fig. 11). In this regard, they proposed Positional Encoding Generators (PEGs) which utilized depth-wise convolutions to make positional embeddings more local and translational equivalent. They also developed a ViT based on the proposed scheme that utilized their PEGs to incorporate more positional information into their architecture and showed good results. In addition, they also showed that instead of the class token, the global average pooling layer above the final MLP layer resulted in boosted performance. Xiao et al. in their study estimated that utilizing CNN layers at early layers in the ViTs can increase their performance (Xiao et al. 2021). For comparison, they replaced the conventional ViT patching with a convolutional stem and reported more generalized and enhanced performance.

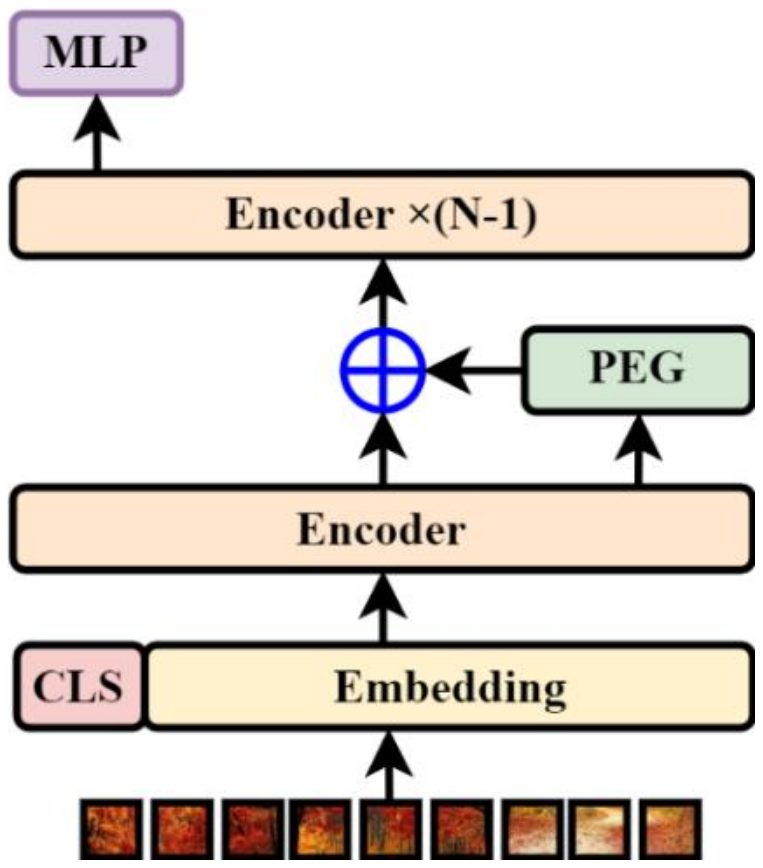

*Figure 11: Architecture of CPVT, which incorporated CNN in their PEG block.*

### 3.6.2. Lateral-layer integration

Models that use a CNN layer or block at the end of the transformer network, such as in place of the last linear layer, or as a post-processing layer fall under this category.

***Dense Prediction Transformer (DPT)***

Ranftl et al. proposed a dense prediction transformer "DPT" for segmentation in natural images. DPT has an encoder-decoder-based design, with a ViT as the encoder and a CNN as the decoder. It captured the global perspective and long-range dependencies by the backbone architecture. The learned global representations were then decoded into image-based embeddings taken by utilizing a CNN. Outputs from the ViT-based encoder were decoded at different levels to carry out dense predictions (Ranftl et al. 2021).

***Local Vision Transformer (LocalViT)***

Li et al, in their research, also incorporated locality into ViT architecture for image classification. The architecture of LocalViT is just like a conventional ViT, with its MSA module specialized to capture global-level features of images. The feed-forward network in ViT encoders performs final predictions by taking input from the learned encodings from the attention module. LocalVit modifies its FFN to incorporate local information into its architecture by employing depth-wise convolution (Li et al. 2021c).

### 3.6.3. Sequential integration

This category describes some of the popular hybrid ViTs that leveraged the benefits of CNN in their ViT architectures by following some sequential integration (Wang et al. 2023c).

### *Convolution and Attention Networks (CoAtNet)*

Dai et al. carried out an extensive study to find out the most optimal and efficient way of merging convolutions and attention mechanisms in a single architecture to increase its generalization and capacity (Dai et al. 2021). In this regard, they introduced CoAtNet, by vertically stacking several convolutional and transformer blocks. For the convolutional blocks, they employed MBConv blocks which are based on depth-wise convolutions. Their findings suggested that stacking two convolutional blocks followed by two transformers blocks, sequentially showed efficient results.

### *CNNs Meet Transformers (CMT)*

Despite their successful performance, ViTs face three main issues, a) inability to capture low-level features by considering correlation in the local neighborhood, b) expensive in terms of computation and memory consumption due to their MSA mechanism, c) and fixed sized input tokens, embedding. To overcome these issues, there comes the boom of hybridization of CNNs and ViTs after 2021. Guo et al. in 2021 also proposed a hybrid ViT, named CMT (CNNs Meet Transformers) (Guo et al. 2021). Inspired by CNNs (Tan and Le 2019), CMT also consists of an initial stem block followed by the sequential stacking of the CNN layer and CMT block. The designed CMT block was inspired by the ViT architecture, therefore contained a lightweight MSA block in place of conventional MSA, and the MLP layer was replaced with an inverted residual feed-forward network (IRFFN). In addition, a local perception unit (LPU) is added to the CMT block to increase the representation capacity of the network. The architecture is shown in Fig. 12.

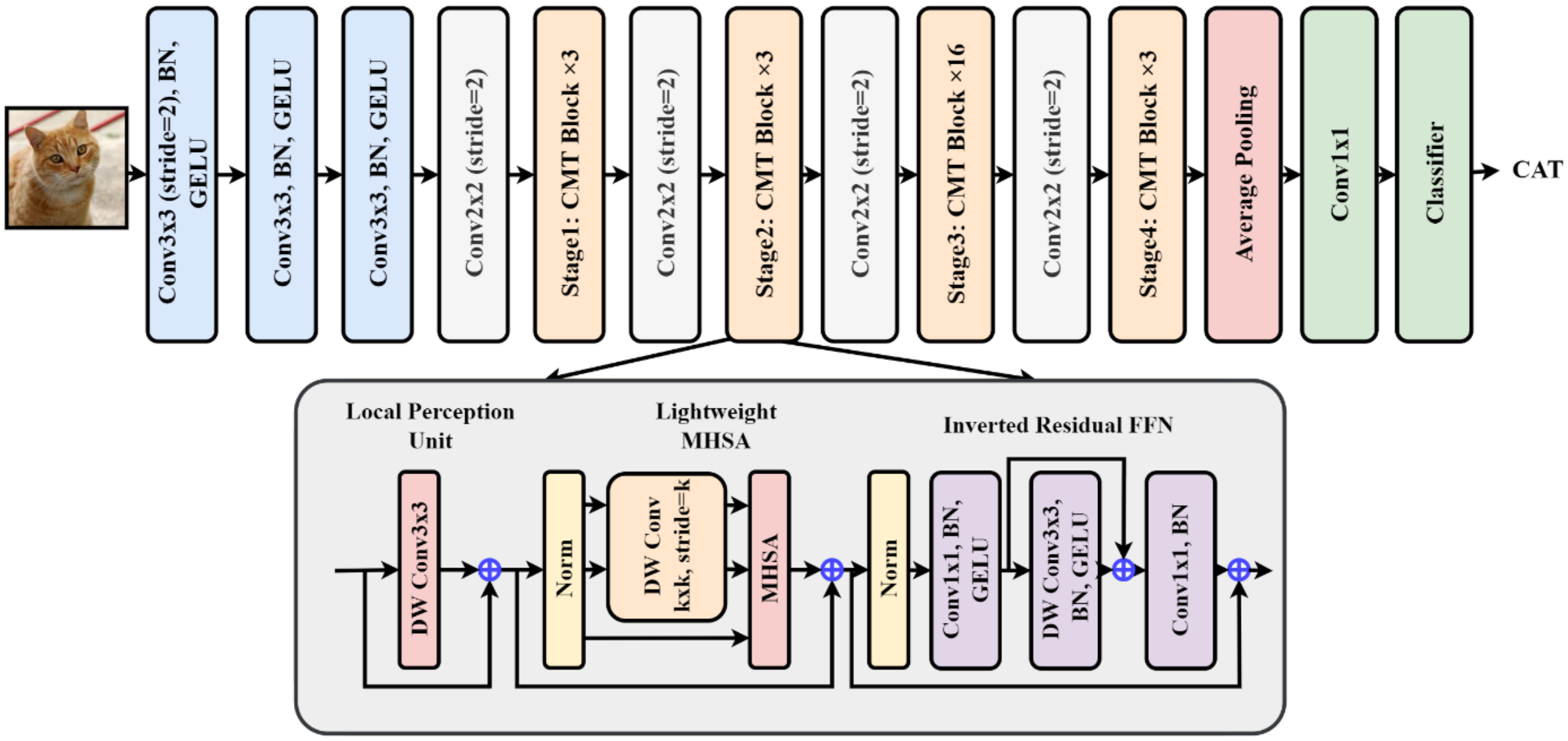

*Figure 12: Architecture of CMT, with the integration of CNN in sequential order*

***Bottleneck Transformers (BoTNet)***

Since the convolutional layer captures low-level features that are the main building blocks of many structuring elements in the images, Srinivas et al, introduced a hybrid ViT, BoTNet (Bottleneck Transformers for Visual Recognition) to benefit both from the CNN and ViT (Srinivas et al. 2021). The architecture of BoTNet is simply a sequential combination of ResNet blocks where the attention mechanism is incorporated in the last three blocks. ResNet block contains two 1x1 convolutions and a 3x3 convolution. The MSA is added in place of 3x3 convolution to capture long-term dependencies in addition to local features.

### 3.6.4. Parallel integration

This category includes those HVT architectures that use both CNNs and transformer architectures in parallel and their predictions are then combined in the end (Wang et al. 2021a).

***Convolution-augmented Transformer (Conformer)***

In 2021, Peng et al. conducted a study to perform visual recognition in natural images. In this regard, they proposed an architecture named, Conformer (Peng et al. 2021). Due to the popularity of ViTs the architecture of Conformer was also based on ViTs. To improve the perception capacity of the network, they integrated the benefits of CNN and to multi-head self-attention mechanism. Conformer, a hybrid ViT, contained two separate branches, a CNN branch to capture local perceptions and a transformer branch to capture global-level features. Subsequent connections were built from the CNN branch to the transformer branch to make each branch local-global context-aware. Final predictions were obtained from a CNN classifier and a transformer classifier. Cross-entropy loss function was used to train each classifier. Conformer showed better performance than other outperforming ViT architectures such as DeiT, and VIT.

***MobileNet-based Transformer (Mobile-Former)***

Chen et al. proposed a concurrent hybrid ViT architecture with two different pathways for a CNN and transformer (Chen et al. 2022e). Like other hybrid ViTs, Mobile-Former employed the CNN model to learn spatial correlation and used a transformer to capture long-term dependencies in the images, thus fusing both the local correlation and global representations. The CNN architecture was based on MobileNet, which uses inverted residual blocks with a reduced number of parameters. Information among both the branches was synchronized using connections, which kept the CNN pathway aware of global information and the transformer aware of local information. Concatenated output from both branches followed by a pooling layer was then fed to a two-layer classifier for final predictions. Fig. 13 shows their detailed architecture.

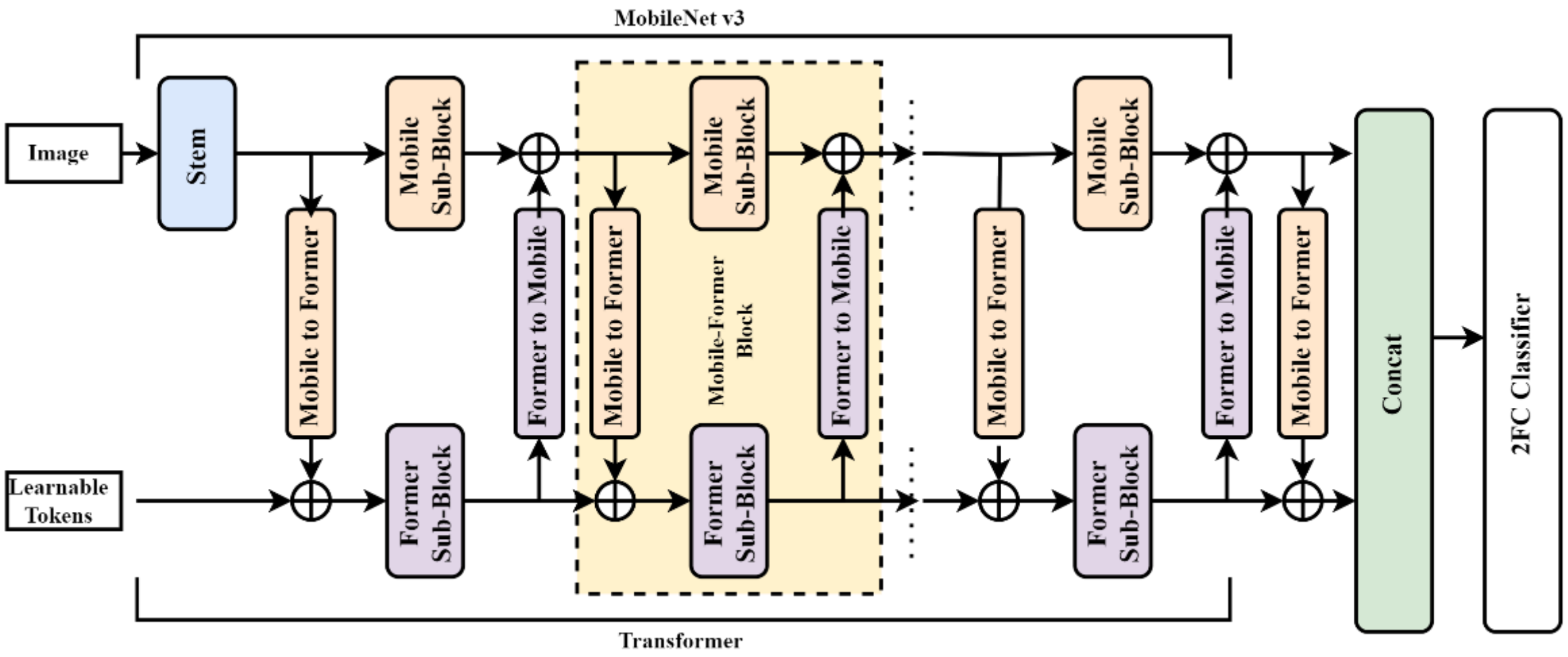

*Figure 13: Architecture of Mobile-former (CNN and transformer with parallel integration)*

***Block-wisely Self-supervised Neural Architecture Search (BossNAS)***

Li et al. developed a search space (HyTra) to evaluate hybrid architectures and advised that each block should be trained separately (Li et al. 2021a). In every layer within the HyTra search space, they utilized CNN and transformer blocks with various resolutions in parallel and freely selectable form. This broad search area includes conventional CNNs with progressively smaller spatial scales and pure transformers with fixed content lengths.

### 3.6.5. Hierarchical integration

Those HVT architectures that adopt a hierarchical design, similar to CNNs, fall under this category. Many of these models have designed a unified block for integrating CNN and ViT, which is then repeated throughout the architecture (Tu et al. 2022b).

***Multi-Axis Attention-based Vision Transformer (MaxViT)***

MaxViT is a variant of the ViT architecture that was introduced by Tu et al., in their paper "Multi-Axis Attention Based Vision Transformer" (Tu et al. 2022b). It introduced the Multi-Axis attention mechanism consisting of blocked local and dilated global attention. It proved to be an efficient and scalable attention mechanism as compared to previous architectures. A new hybrid block was introduced as a basic element, which consists of MBConv-based convolution and Multi-Axis based attention. The basic hybrid block was repeated over multiple stages to obtain a hierarchical backbone, similar to CNN-based backbones that can be used for classification, object detection, segmentation, and generative modeling. MaxViT can see locally and globally across the whole network, including the earlier stages.

### *Convolutional Vision Transformer (CvT)*

CvT was introduced by Wu et al. in 2021 (Wu et al. 2021a). The architecture of CvT contained several stages like CNNs to make up a hierarchical framework. They added convolution in their architecture in two ways. At first, they used a convolutional token embedding to extract token sequences, which not only incorporated locality in the network but also shortened the sequence length gradually. Secondly, they proposed a convolutional projection that used depth-wise separable convolution to replace the linear projection before each self-attention block in the encoder block. CvT outperformed other approaches for image recognition.

### *Vision-Friendly Transformer (Visformer)*

Visformer was introduced as a vision-friendly transformer in 2020 (Chen et al. 2021d) presenting a modular design for efficient performance. The architecture had several modifications to a conventional ViT network. In Visformer, global average pooling was employed in place of classification token, and layer normalization was replaced with batch normalization. In addition, they utilized convolutional blocks inspired by ResNeXt (Xie et al.) instead of self-attention in each stage to efficiently capture both spatial and local features. However, to model the global dependencies they adopted self-attention in the last two stages. Another notable modification in Visformer's architecture was the addition of 3x3 convolutions in the MLP block.

### *Vision Transformer Advanced by Exploring intrinsic Inductive Bias (ViTAE)*

The authors suggested a novel ViT architecture called ViTAE, that combines two different basic cell types (shown in Fig. 14): reduction cells (RC) and normal cells (NC) (Xu et al. 2021b). RCs are used to downscale input images and embed them into enriched multi-scale contextual tokens, while NCs are used to model local and long-term dependencies concurrently within the token sequence. The underlying structure of these two types of cells is also similar, consisting of parallel attention modules, convolutional layers, and an FFN. RC includes contextual information in tokens by utilizing several convolutions with different dilation rates in the pyramid reduction module. The authors also presented a more optimized version, ViTAEv2, that showed better performance than earlier method (Zhang et al. 2022d).

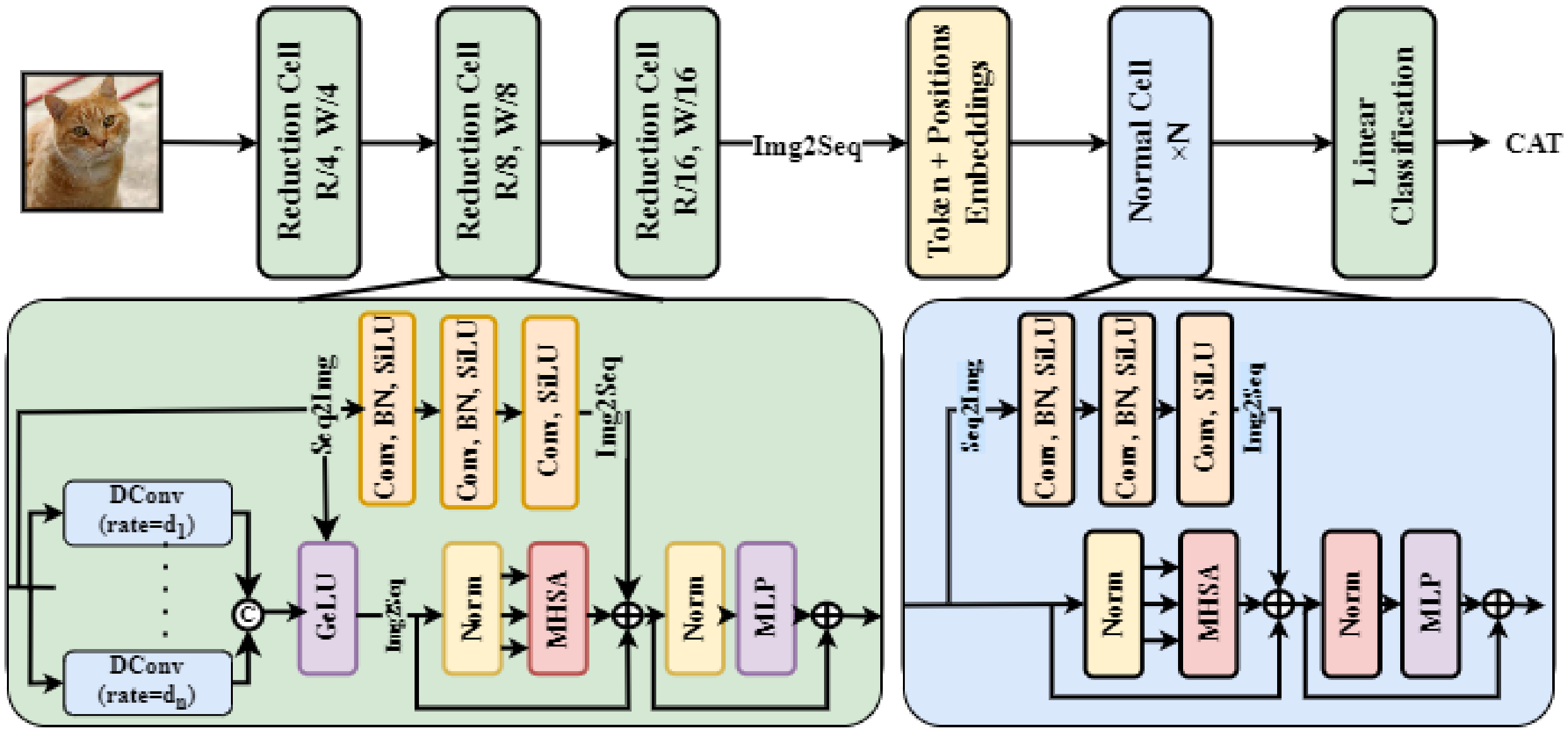

*Figure 14: Architectural diagram of ViTaE*

***Convolution-Transformer Network (ConTNet)***

A novel Convolution-Transformer Network (ConTNet) is proposed for computer vision tasks to address the challenges faced in this area. The ConTNet is implemented by stacking multiple ConT blocks (Yan et al.) (shown in Fig. 15). The ConT block treats the standard transformer encoder (STE) as an independent component similar to a convolutional layer. Specifically, a feature map is divided into several patches of equal size and each patch is flattened to a (super) pixel sequence, which is then input to the STE. After reshaping the patch embeddings, the resulting feature maps are then passed on to the next convolutional layer or to the STE module.

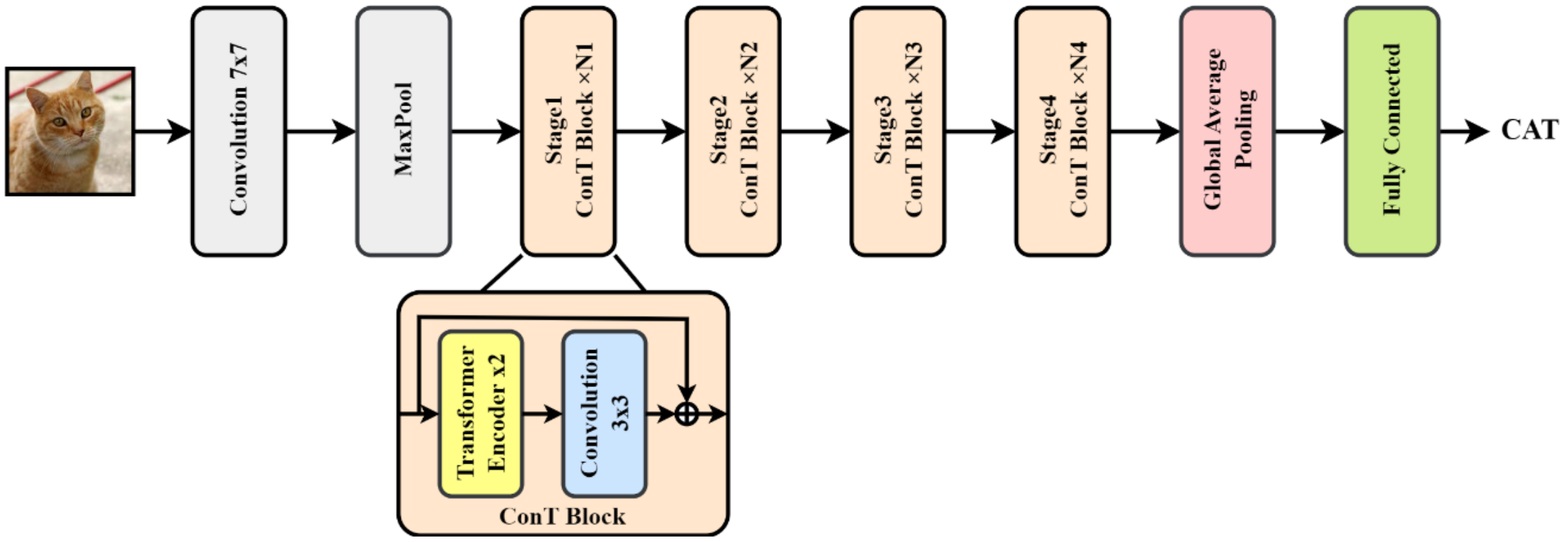

*Figure 15: Architecture of ConTNet, which integrated CNN and ViT in its ConT block to form a hierarchical architecture.*

### 3.6.6. Attention-based integration

This section discusses those HVT architectures, which have utilized CNNs in their attention mechanism to incorporate locality.

***Evolving Attention with Residual Convolutions (EA-AA-ResNet)***

Due to the limited generalizability of independent self-attention layers in capturing underlying dependencies between tokens, Wang et al. extended the attention mechanism by adding convolutional modules (Wang et al.). Specifically, they adopted a convolutional unit with residual connections to generalize the attention maps in each layer by exploiting the knowledge inherited from previous layers, named as Evolving Attention (EA). The proposed EA-AA-ResNet architecture extends attention mechanisms by bridging attention maps across different layers and learning general patterns of attention using convolutional modules.

**ResNet Transformer (ResT)**

A hybrid architecture that integrates convolution operation in its attention mechanism, allowing it to capture both global and local features effectively (Zhang and Yang 2021). The authors utilized a new efficient transformer block in their architecture where they replaced the conventional MSA block with its efficient variant. In the proposed efficient multi-head self-attention, they employed depth-wise convolution to reduce the spatial dimensions of the input token map before computing the attention function.

***Convolution-Enhanced Image Transformer (CeiT)***

CeiT was proposed by Yuan et al. in 2021 in their paper "Incorporating Convolution Designs into Visual Transformers" (Yuan et al. 2021a). The proposed CeiT combined the benefits of CNNs and ViTs in extracting low-level features, capturing locality, and learning long-range dependencies. In their CeiT, they made three main advancements in conventional ViT architecture. They modified the patch extraction scheme, the MLP layer and added a last layer above the ViT architecture. For patch extraction, they proposed an Image-to-Tokens (I2T) module in which they utilized CNN-based blocks to process the inputs. Instead of utilizing raw input images, they used low-level features learned from the initial convolutional blocks to extract patches. I2T contained convolutional, max pooling, and batch normalization layers in its architecture to fully leverage the benefits of CNNs in ViTs. They utilized a Locally-enhanced Feed-Forward (LeFF) layer in place of the conventional MLP layer in the ViT encoder, in which depth-wise convolutions were utilized to capture more spatial correlation. In addition, a last class token attention (LCA) layer was devised to systematically combine the outputs from different layers of ViT. CeiT not only showed promising results on several image and scene recognition datasets (including ImageNet, CIFAR, and Oxford-102) but is also computationally efficient as compared to ViT.

### 3.6.7. Channel boosting-based integration

Channel boosting (CB) is an idea used in DL to increase the representation learning ability of CNN models. In CB, besides the original channels, boosted channels are generated using transfer learning-based auxiliary learners to capture diverse and complex patterns from images. CB-based CNNs (CB-CNN) have shown outstanding performance in various vision-related tasks. In a study by Ali et al., they proposed a CB-based HVT architecture (Ali et al. 2023b). In CB-HVT they utilized CNNs and ViT-based auxiliary learners to generate boosted channels. The CNN-based channels captured local-level diversity in the image patterns, whereas Pyramid Vision Transformer (PVT)-based channels learned global-level contextual information. The authors evaluated CB-HVT on the lymphocyte assessment dataset, where it showed reasonable performance. An overview of their architecture is shown in Fig. 16.

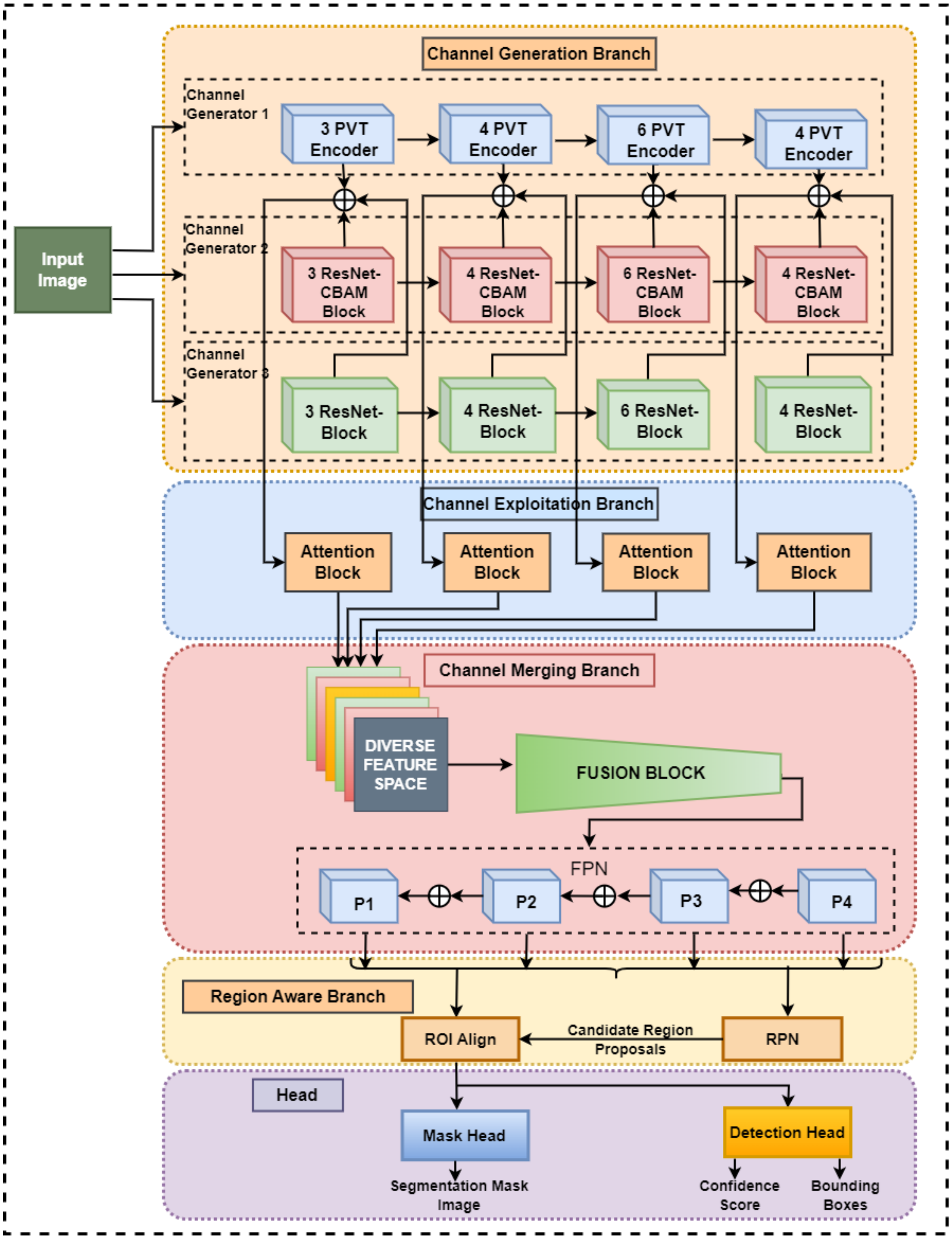

*Figure 16: Overview of CB-HVT, where PVT (a VIT) is combined within CNN architecture using channel boosting.*

### 3.7. Empirical comparison of different methods

In this section, we present a brief yet comprehensive empirical comparison of several ViT and HVT architectures that have demonstrated exceptional performance across various computer vision tasks. To get insights into their strengths and weaknesses, we have provided a detailed overview in Table 3 and Table 4. In addition, we have also highlighted the primary modifications made in each model, along with the underlying rationale, as per their taxonomy.

*Table 3: Empirical comparison of various ViT architectures, based on their strengths, weaknesses, rationale, and performance on benchmark datasets (For comparison, we have reported the results of the best performing variants of the mentioned architecture).*

| Architecture | Strength (Architectural modification) | Weakness | Rationale | Metric |
|---|---|---|---|---|
| **Patch-Based Approaches** | | | | |
| **T2T-ViT** (Yuan et al. 2021b) CODE | Utilized Token-to-Token module for iterative patching. | May have higher computational requirements due to increased tokenization. | To improve the representation by focusing on tokens instead of patches. | **ImageNet** 83.3% Top-1 Acc **@ 384×384** |
| **TNT-ViT** (Han et al. 2021) CODE | Utilizes multi-level patching, to capture objects with spatial and size variations. | May require more parameters, leading to increased model size. | To capture the attetnion inside the local patches. | **ImageNet** 82.9% Top-1 Acc **@ 224x224** |
| **DPT** (Chen et al. 2021e) CODE | Used DePatch, to have patches of variable sizes. | Could be sensitive to the selection of deformable patches, impacting performance. | For better handling of irregular-shaped objects in the image. | **ImageNet** 81.9% Top-1 Acc @ 224x224 |
| **CrowdFormer** (Yang et al. 2022b) | Models global context by learning features at different scales, effective for crowd counting. | Could be computationally expensive for real-world applications. | The global context is incorporated to deal with the uneven distribution of crowds. | **NWPU** 67.1 MAE, 301.6 MSE **@ 512x512** |
| **Knowledge Transfer-Based Approaches** | | | | |
| **DeiT** (Touvron et al. 2020) CODE | Based on a teacher-student model, where knowledge from teacher is transferred to student model | Performance is highly dependent upon the transferred knowledge. | Enables efficient image representation, suitable for resource-constrained scenarios. | **ImageNet** 85.2% Top-1 Acc **@ 384x384** |
| **TaT** (Lin et al. 2022) CODE | Utilized one-to-all spatial. matching knowledge distillation approach | Relatively computationally expensive due to one-to-many mapping | For an effective knowledge transfer one-to-many mapping was done between the teacher and the student models. | **ImageNet** 72.41% Top-1 Acc **@ 513x513** |
| **TinyViT** (Wu et al. 2022a) CODE | Compact model with reduced output logits of teacher model to have a light model. | May not achieve the same level of accuracy as larger, more complex models. | Provides an efficient and lightweight model, efficient for deployment. | **ImageNet** 86.5% Top-1 Acc **@ 512x512** |
| **Shifted Window-Based Approaches** | | | | |
| **Swin Transformer** (Liu et al. 2021c) CODE | Shifted windowing scheme allows to efficiently compute self-attention over large images. | Relatively computationally expensive, which may limit its applicability to real-time applications. | To enable cross-window connections, which can help in enhancing the learning capacity of the model. | **ImageNet** 84.5% Top-1 Acc **@ 384x384** |
| **Attention-Based Approaches** | | | | |
| **CaiT** (Touvron et al. 2021) CODE | LayerScale and class-attention stage significantly improve the accuracy of deep transformers. | Performance may be influenced by class attention accuracy and distribution. | To give more focus to important classes, leading to better classification. | **ImageNet** 86.5% Top-1 Acc **@ 448x448** |
| **DAT** (Xia et al. 2022) CODE | Utilizes deformable attention mechanisms, improving feature representation. | The deformable attention may have increased computational cost. | To enable better modeling of local image structures and deformable objects. | **ImageNet** 84.8% Top-1 Acc **@ 384x384** |
| **SeT** (Sun et al. 2022) | Factorizes the spatial attention into pixel- | May have limited representation | To effectively capture both fine-grained and | **ImageNet** 83.3% |
| | wise and patch-wise attention, which reduces the computational cost. | capacity compared to non-separable approaches. | coarse-grained features. | Top-1 Acc **@ 224x224** |
| **Multi-Transformer-Based Approaches** | | | | |
| **CrossViT** (Chen et al. 2021a) CODE | Capable of modeling multi-scale feature representations in linear time. | Does not explore the use of different token fusion strategies. | To explore multi-scale tokens and demonstrate the effect of dual-branch feature fusion | **ImageNet** 82.8% Top-1 Acc **@ 224x224** |
| **Dual-ViT** (Yao et al.) CODE | The semantic pathway captures global semantics and a pixel pathway learns finer local details. | Complex architecture as compared to other SOTA Transformers which could make it difficult to train. | To capture global and local features simultaneously and effectively. | **ImageNet** 85.7% Top-1 Acc **@ 384x384** |
| **MMViT** (Liu et al. 2023b) CODE | Combines multi-scale and multi-view images, improving performance and robustness. | May require additional pre-processing steps to handle multi-view images. | To enable comprehensive image understanding by leveraging multiple scales and views. | **ImageNet** 83.2% Top-1 Acc **@ 224x224** |
| **MPViT** (Lee et al. 2021b) CODE | Multi-scale patch embedding and multi-path structure helps in learning fine and coarse feature representations. | Increased depth and complexity may lead to higher training time and resource usage. | To demonstrate that multi-scale and multi-path structures are beneficial for dense prediction tasks. | **ImageNet** 83.4% Top-1 Acc **@ 384x384** |

*Table 4: Empirical comparison of several HVT architectures, based on their strengths, weaknesses, rationale and performance on benchmark datasets (For comparison, we have reported the results of the best performing variants of the mentioned architecture)*

| Architecture | Strength | Weakness | Rationale | Metric |
|---|---|---|---|---|
| **Early-Layer Integration** | | | | |
| **DETR** (Carion et al. 2020) CODE | End-to-end object detection with transformers, eliminating the need for separate region proposal networks. | The approach maybe computationally expensive, especially for large images. | To demonstrate the effectiveness of transformers in dense prediction tasks. | **COCO** 44.9% Box mAP @ **384x384** |
| **LeViT** (Graham et al. 2021) CODE | Utilized convolution layers in initial layers to capture local features and reduce image resolutions. | May require more memory than some of the best CNNs. | To combine the strengths of both CNNs and ViTs, while avoiding some of their weaknesses. | **ImageNet** 82.6% Top-1 Acc @ **384x384** |
| **CPVT** (Chu et al. 2021b) CODE | Utilized a new scheme for conditional position encodings for boosted performance | May have increased complexity and computational requirements. | To make positional embeddings translational invariant, they used depth-wise convolutions | **ImageNet** 82.7% Top-1 Acc @ **224x224** |
| **Lateral-Layer Integration** | | | | |
| **DPT** (Ranftl et al. 2021) CODE | Utilized a ViT-based and a CNN-based decoder | Higher memory requirements compared to some traditional dense prediction models. | To demonstrate the suitability of transformers for dense prediction tasks by capturing long-range dependencies between pixels. | **ADE20K** 49.02% IoU @ **520x520** |
| **LocalViT** (Li et al. 2021c) CODE | Introduces depth-wise convolutions into its FFN. | May have limitations in capturing global context information effectively. | To explicitly incorporate local dependencies. | **ImageNet** 80.8% Top-1 Acc @ **224x224** |
| **Sequential Integration** | | | | |
| **CoAtNet** (Dai et al. 2021) CODE | Integrates convolutional layers within MSA. | Not as accurate as other ViTs on large datasets. | To effectively combine the strengths of both convolutional and MSA mechanism | **ImageNet** 86.0% Top-1 Acc @ **512x512** |
| **CMT** (Guo et al. 2021) CODE | Incorporated a lightweight MSA | Complex than traditional CNNs, difficult to implement and train. | To integrate the strengths of both CNNs and Transformers for vision tasks while avoiding some weaknesses. | **ImageNet** 84.8% Top-1 Acc @ **288x288** |
| **BoTNet** (Srinivas et al. 2021) | Employs bottleneck structures to improve memory efficiency and computational speed. | Performance may be influenced by the choice of bottleneck design and model depth. | To enhance the efficiency and speed of ViTs through bottleneck structures. | **ImageNet** 84.7% Top-1 Acc @ **224x224** |
| | | **Parallel Integration** | | |
| **Conformer** (Peng et al. 2021) CODE | Feature Coupling Unit (FCU) allows for efficient fusion of local features and global representations. | As compared to traditional CNNs, it may have complex to train and deploy. | To combine the strengths of both CNNs and self-attention mechanisms, while mitigating their weaknesses. | **ImageNet** 84.1% Top-1 Acc @ **224x224** |
| **Mobile-Former** (Chen et al. 2022e) CODE | The bridge between MobileNet and transformer enables bidirectional fusion of local and global features. | The light-weight cross attention in the bridge may not be able to fully capture the interactions between local and global features. | To provide parallel interaction of MobileNet and transformer, allowing the model to achieve a good balance between efficiency and representation power. | **ImageNet** 79.3% Top-1 Acc @ **224x224** |
| **BossNAS** (Li et al. 2021a) CODE | Can effectively search for hybrid CNN-transformer architectures. | Can be computationally expensive to train, especially for large search spaces. | Large and diverse search space of hybrid architectures makes it difficult for traditional NAS methods to be effective. | **ImageNet** 82.5% Top-1 Acc @ **512x512** |
| | | **Hierarchical Integration** | | |
| **MaxViT** (Tu et al. 2022b) CODE | Introduces a number of novel ideas, including multi-axis attention, hierarchical stacking, and linear-complexity global attention. | Can be more difficult to train because of the complex attention mechanism and may require more data to achieve good results. | To enable local and global feature extraction through self-attention in linear time. | **ImageNet** 86.70% Top-1 Acc @ **512x512** |
| **CvT** (Wu et al. 2021a) CODE | Combines convolutional and MSA blocks, striking a balance between efficiency and accuracy. | Performance may be influenced by the specific configuration of the CvT architecture. | Integrates convolutional and ViT elements for effective vision tasks. | **ImageNet** 87.7% Top-1 Acc @ **384x384** |
| **Visformer** (Chen et al. 2021d) CODE | Optimizes transformer architecture for vision tasks, considering image-specific challenges. | May require further architectural advancements to achieve state-of-the-art performance. | To tailor the transformer architecture for vision-specific challenges. | **ImageNet** 82.19% Top-1 Acc @ **224x224** |
| **ViTAE** (Xu et al. 2021b) CODE | Introduces an inductive bias-aware architecture, improving generalization. | May not be as effective for tasks that require fine-grained visual reasoning | To incorporate the inductive biases to enhance model generalization and adaptability. | **ImageNet** 83.0% Top-1 Acc @ **384x384** |
| **ConTNet** (Yan et al.) CODE | More robust to changes in the input data than transformer-based models. | Model complexity may increase due to the combination of convolution and transformers. | To obtain hierarchical features using both convolution and transformers for various vision-related tasks. | **ImageNet** 81.8% Top-1 Acc @ **224x224** |
| **Attention-Based Integration** | | | | |
| **EA-AA-ResNet** (Wang et al.) CODE | Evolves attention mechanisms with residual convolutions, enhancing feature representation. | May have higher computational cost compared to standard convolutional models. | To improve feature representation through evolving attention with residual convolutions. | **ImageNet** 79.63% Top-1 Acc @ **224x224** |
| **ResT** (Zhang and Yang 2021) CODE | The introduction of Memory-Efficient MSA, Spatial Attention for positional encoding and Stack of conv. layers for patch embedding. | May be computationally more expensive than traditional CNN-based models. | To provide novel and innovative techniques that make transformer models efficient and versatile for visual recognition tasks. | **ImageNet** 83.6% Top-1 Acc @ **224x224** |
| **CeiT** (Yuan et al. 2021a) CODE | Enhances ViTs with convolutional operations, improving efficiency and performance. | Model complexity may increase with the addition of convolutional operations. | To improve local features extraction of ViTs with convolutional components. | **ImageNet** 83.3% Top-1 Acc @ **384x384** |
| **Channel Boosting-Based Integration** | | | | |
| **CB-HVT** (Ali et al. 2023b) | Employs channel boosting for better feature representation, improving model accuracy. | Increased computational cost due to additional channel boosting computations. | To enhance feature representation through channel boosting in a hybrid architecture. | **LYSTO** 88.0% F-Score @ **256x256**<br>**NuClick** 82.0% F-Score @ **256x256** |

# 4. Applications of ViTs and HVTs

In recent years, both ViTs and HVTs have gained prominence across a range of vision-based applications (Lee et al. 2021b; Deng et al. 2023; Xue and Ma 2023; Yan et al. 2023; Cheng et al. 2023; Lian et al. 2023; Chen et al. 2023d; Dehghani et al. 2023; Liu et al. 2023a; Ye et al. 2023a), including image and video recognition (Pecoraro et al. 2022; Zhang and Zhang 2022; Chen et al. 2023a; Xia et al. 2023; Mogan et al. 2023; Liang et al. 2023), object detection (Wang et al. 2023c) (Dehghani-Dehcheshmeh et al. 2023; Huang et al. 2023a; Yu and Zhou 2023), segmentation (Chen et al. 2021b; Li et al. 2023e; Quan et al. 2023), image restoration (Zhou et al. 2023a), and medical image analysis (Gao et al. 2021; An et al. 2022; Chen et al. 2022a; Song et al. 2022a; Zhang et al. 2022b; Yang and Yang 2023; Nafisah et al. 2023; Wu et al. 2023c). Transformer-based modules and CNNs are combined to create HVTs, an effective approach that can interpret intricate visual patterns (Li et al. 2023a). Some notable applications for ViTs and HVTs are discussed below.

## 4.1. Image/video recognition

CNNs have been extensively utilized for image and video processing due to their capability to automatically extract complex information from visual data (Fan et al. 2016; Fang et al. 2018; Yao et al. 2019; Kaur et al. 2022; Rafiq et al. 2023). Nevertheless, ViTs have revolutionized the field of computer vision by achieving outstanding performance on various challenging tasks, including image and video recognition (Chen and Ho 2022; Jing and Wang 2022; Chen et al. 2022c, b; Wensel et al. 2022; Ulhaq et al. 2022). The success of ViTs can be attributed to their self-attention mechanism, which enables them to capture long-range dependence in images (Ji et al. 2023). In recent years, HVTs have gained popularity, as they combine the power of both CNNs and transformers (Zhang et al. 2022a; Li et al. 2023g; Ma et al. 2023a). Various methods have been proposed based on HVTs for recognition in both images and videos (Jiang et al. 2019; Li et al. 2021b; Huang et al. 2021a; GE et al. 2021; Yang et al. 2022a; Leong et al. 2022; Zhao et al. 2022a; Raghavendra et al. 2023; Zhu et al. 2023b). Xiong et al. proposed a hybrid multi-modal approach based on ViT and CNN to enhance fine-grained 3D object recognition (Xiong and Kasaei 2022). Their approach encodes the global information of the object using the ViT network and the local representation of the object using a CNN network through both RGB and depth views of the object. Their technique outperforms both CNN-only and ViT-only baselines. In another technique, Tiong et al. presented a novel hybrid attention vision transformer (HA-ViT) to carry out face-periocular cross identification (Tiong et al. 2023). HA-ViT utilizes depth-wise convolution and convolution-based MSA concurrently in its hybrid attention module in parallel to integrate local and global features. The proposed methodology outperforms three benchmark datasets in terms of Face Periocular Cross Identification (FPCI) accuracy. Wang et al. proposed a novel approach for visual place recognition using an HVT-based architecture (Wang et al. 2022d). Their method aims to improve the robustness of the visual place recognition system by combining both CNN and ViT to capture local details, spatial context, and high-level semantic information. To recognize vehicles Shi et al. developed a fused network that used SE-CNN architecture for feature extraction followed by the ViT architecture to capture global contextual information (Shi et al. 2023). Their proposed approach demonstrated good accuracy values for the road recognition task.

## 4.2. Image generation

Image generation is an interesting task in computer vision and can serve as a baseline for many downstream tasks (Frolov et al. 2021). Generative adversarial networks (GANs) are widely used for image generation in various domains (Arjovsky et al. 2017; Karras et al. 2019). Additionally, ViT-based GANs have shown promising performance in this task (Lee et al. 2021a; Naveen et al.

2021; Rao et al. 2022; Gao et al. 2022b). Recently, researchers have also utilized HVT-based GANs and demonstrated outstanding performance on various benchmark datasets (Tu et al. 2022a; Lyu et al. 2023). Torbunov, et al. reported UVCGAN, a hybrid GAN model, for image generation (Torbunov et al. 2022). The architecture of the UVCGAN model is based on the original CycleGAN model (Zhu et al. 2017) with some modifications. The generator of UVCGAN is a hybrid architecture based on a UNet (Weng and Zhu 2015) and a ViT bottleneck (Devlin et al. 2018). Experimental results demonstrated its superior performance compared to earlier best performing models while retaining a strong correlation between the original and generated images. In another work, SwinGAN was introduced for MRI reconstruction by Zhao, et al (Zhao et al. 2023). They utilized Swin Transformer U-Net-based generator network and CNN-based discriminator network. The generated MRI images by SwinGAN showed good reconstruction quality due to its ability to capture more effective information. Tu et al. combined the Swin transformer and CNN layers in their proposed SWCGAN (Tu et al. 2022a). In their architecture they utilized CNN layers initially to capture local level features and then in later layers utilized Residual Dense Swin Transformer Blocks "RDST" to capture global level features. The developed method showed good reconstruction performance compared to existing approaches in remote sensing images. Recently, Bao et al. proposed a spatial attention-guided CNN-Transformer aggregation network (SCTANet) to reconstruct facial images (Bao et al. 2023b). They utilized both CNN and transformer in their Hybrid Attention Aggregation (HAA) block for deep feature extraction. Their experimental results demonstrated better performance than other techniques. Zheng et al. in their approach presented a HVT-based GAN network for medical image generation (Zheng et al. 2023). In their approach, named L-former they utilize transformers in the shallow layers and CNNs in the deeper layers. Their approach demonstrated outperformance as compared to conventional GAN architectures.

### 4.3. Image segmentation

Although CNNs and ViT-based approaches have shown exceptional performance in complex image-related tasks such as image segmentation, there is currently an emphasis on combining the strengths of both approaches to achieve boosted performance (Dolz et al. 2019; Wang et al. 2020, 2022c, b; Jing et al. 2023; Shafri et al. 2023; Yang et al. 2023c). In this regard, Wang et al. presented a new semantic segmentation method called DualSeg for grapes segmentation (Wang et al. 2023a). Their method combines Swin Transformer and CNN to leverage the advantages of both global and local features. In another work, Zhou and co-authors proposed a hybrid approach named SCDeepLab to segment tunnel cracks (Zhou et al. 2023b). Their approach outperformed other CNN-only and transformer-only-based models in segmenting cracks in tunnel lining. Feng et al. carried out segmentation recognition in metal couplers to detect fracture surfaces (Feng et al. 2023). To this end, they proposed an end-to-end HVT-based approach by utilizing a CNN for automatic feature extraction and a hybrid convolution and transformer (HCT) module for feature fusion and global modeling. Recently, Xia and Kim developed Mask2Former, an HVT approach, to address the limitations of ViT or CNN-based systems (Xia and Kim 2023). The developed approach achieved better results as compared to other techniques on both the ADE20K and Cityscapes datasets. Li et al. proposed an HVT-based method called MCAFNet for semantic segmentation of remote sensing images (Li et al. 2023d).

### 4.4. Image Restoration

A crucial task in computer vision is image restoration, which tends to restore the original image from its corrupted version. Image restoration-based systems have generally shifted from the use of CNNs to ViT models (Ali et al. 2023a; Song et al. 2023) and more recently to HVTs that combine the strengths of both CNNs and transformers (Gao et al. 2022a; Wu et al. 2023c). Yi et al. proposed an auto-encoder based hybrid method to carry out single infrared image blind deblurring (Yi et al. 2023). Their approach utilizes hybrid convolution-transformer blocks for extracting context-related information between the objects and their backgrounds. To hasten the convergence of the training process and achieve superior image deblurring outcomes, the study also incorporated a multi-stage training technique and mixed error function. In another technique, Chen et al. developed an efficient image restoration architecture called Dual-former, which combines the local modeling ability of convolutions and the global modeling ability of self-attention modules (Chen et al. 2022d). The proposed architecture achieves superior performance on multiple image restoration tasks while consuming significantly fewer GFLOPs than previously presented methods. To address the issue of high computational complexity Fang et al. utilized a hybrid network, HNCT, for lightweight image super-resolution (Fang et al. 2022). HNCT leverages the advantages of both CNN and ViT and extracts features that consider both local and non-local priors, resulting in a lightweight yet effective model for super-resolution. Experimental results demonstrate that HNCT's improved results as compared to existing approaches with fewer parameters. Zhao et al. developed a hybrid denoising model, called Transformer Encoder and Convolutional Decoder Network (TECDNet), for efficient and effective real image denoising (Zhao et al. 2022b). TECDNet attained outstanding denoising results while maintaining relatively a low computational cost. Recently, Chen et al. presented an end-to-end HVT-based image fusion approach for infrared and visible image fusion (Chen et al. 2023b). The proposed technique consists of a CNN module with two branches to extract coarse features, and a ViT module to obtain global and spatial relationships in the image. Their method was able to focus on global information and overcome the flaws of CNN-based methods. In addition, to retain the textural and spatial information a specialized loss function is designed.

### 4.5. Feature extraction

Feature extraction is essential in computer vision to identify and extract relevant visual information from images. Initially, CNNs were generally used for feature extraction in computer vision applications, but now ViTs have gained attention due to their impressive results in image classification as well as other applications like pose estimation, and face recognition (Wang et al. 2023d; Zhu et al. 2023a; Su et al. 2023).

Li and Li, in their work, presented a hybrid approach, ConVit, to merge the advantages of both CNNs and transformers for effective feature extraction to identify crop disease (Li and Li 2022). The experimental results of the developed approach showed good performance in the plant disease identification task. A cascaded approach was proposed by Li et al. for recaptured scene image identification (Li et al. 2023b). In their approach, they initially employed CNN layers to extract local features and later in the deeper layers, they utilized transformer blocks to learn global-level image representations. The high accuracy value of their proposed approach demonstrated its effectiveness in identifying recaptured images. Li and co-authors developed HVT architecture to detect defects in strip steel surfaces. Their approach utilized a CNN module, followed by a patch embedding block and two transformer blocks to extract high-domain relevant features. Their experiments showed good classification performance as compared to existing methods. Recently, Rajani et al. in their approach, proposed an encoder-decoder approach for categorizing different seafloor types. Their developed method is a ViT-based architecture with its MLP block replaced with CNN-based feature extraction module. The modified architecture achieves outstanding results while meeting real-time computational requirements.

## 4.6. Medical image analysis

CNN-based approaches have been frequently employed for analyzing medical images due to their capability to capture diverse and complex patterns (Zafar et al. 2021; Sohail et al. 2021b; Rauf et al. 2023). Nevertheless, driven by the requirement to model image representations at a global level, researchers have found inspiration in employing ViTs within the field of medical image analysis (Obeid et al. 2022; Cao et al. 2023; Zou and Wu 2023; Li et al. 2023c; Zidan et al. 2023; Xiao et al. 2023). Recently, several studies have proposed integrating CNNs and ViTs to capture both local and global image features in medical images, allowing for more comprehensive analysis (Tragakis et al.; Springenberg et al. 2022; Wu et al. 2022c, 2023a; Jiang and Li 2022; Bao et al. 2023a; Dhamija et al. 2023; Huang et al. 2023b; Ke et al. 2023; Yuan et al. 2023a). These hybrid architectures (CNN-transformer) have shown tremendous performance in a number of medical images-related applications (Zhang et al. 2021c; Shen et al. 2022; Rehman and Khan 2023; Li et al. 2023f; Wang et al. 2023b). Tragakis, et al. proposed a novel Fully Convolutional Transformer (FCT) approach to segment medical images (Tragakis et al.). FCT adapted both ViT and CNN in its architecture by combining the ability of CNNs in learning effective image representations with the ability of Transformers to capture long-term dependencies. The developed approach showed outstanding performance on various medical challenge datasets as compared to other existing architectures. In another work, Heidari, et al. proposed HiFormer, an HVT to capture multi-scale feature representations by utilizing a Swin Transformer module and a CNN-based encoder (Heidari et al. 2022). Experimental results demonstrated the effectiveness of HiFormer in segmenting medical images in various benchmark datasets. In their paper, Yang and colleagues presented a novel hybrid approach called TSEDeepLab, which combines convolutional operations with transformer blocks to analyze medical images (Yang et al. 2023a). Specifically, the approach utilizes convolutional layers in the early stages for learning local features, which are then processed by transformer blocks to extract global patterns. Their approach demonstrated exceptional segmentation accuracy and strong generalization performance on multiple medical image segmentation datasets.

### 4.7. Object Detection

Object detection is a crucial computer vision task with a wide range of real-world applications such as surveillance, robotics, crowd counting, and autonomous driving (Liu et al. 2023a). The progress of DL has significantly contributed to the advancements in object detection over the years (Er et al. 2023). ViTs have also shown impressive performance in object detection due to their self-attention mechanism that allows them to capture long-range dependencies between image pixels and identify complex object patterns across the entire image (Carion et al. 2020; Chen et al. 2021c; Wang and Tien 2023; Heo et al. 2023). Recently, there has been a lot of interest in HVTs for combining CNNs with self-attention mechanisms to improve object detection performance (Jin et al. 2021; Maaz et al. 2022; Mathian et al. 2022; Ye et al. 2023b; Zhang et al. 2023b; Lu et al. 2023a; Ullah et al. 2023). Beal et al. proposed an HVT approach named ViT-FRCNN for object detection in natural images. In their approach, they utilized a ViT-based backbone for a Faster R-CNN object detector. ViT-FRCNN showed improved detection results with a better generalization ability (Beal et al. 2020). Chen et al. introduced a single-stage hybrid detector for detection in remote sensing images. their proposed approach, MDCT leveraged both the CNNs and transformers in its architecture and showed better performance as compared to other single-stage detectors (Chen et al. 2023c). Lu et al. developed an HVT-based approach for object detection in unmanned aerial vehicle (UAV) images (Lu et al. 2023b). The proposed approach utilized a transformer-based backbone to extract features with global-level information, which were then fed to FPN for multi-scale feature learning. The proposed method demonstrated good performance as compared to earlier approaches. Yao and his colleagues proposed a fusion network that utilizes individual transformer and CNN-based branches to learn global and local-level features (Yao et al. 2023). Experimental results showed satisfactory performance of the developed method as compared to other methods.

### 4.8. Pose Estimation

Human pose estimation tends to identify important points in various scenarios. Both CNNs and transformers have shown exemplary performance in pose estimation tasks (Sun et al. 2019; Huang et al. 2019; Cao et al. 2022). In a recent study, a method for multi-person pose estimation has been described, utilizing the multi-head attention mechanism of ViT (Yang et al. 2023b). Currently, researchers are focusing to combine CNNs and transformers in a unified method to incorporate both local and global level information for accurate pose estimation (Stoffl et al. 2021; Mao et al. 2021; Li et al. 2021d; Wu et al. 2022b). Zhao et al. presented a new Dual-Pipeline Integrated Transformer "DPIT" for human pose estimation (Zhao et al. 2022c). In Zhao's approach initially, two CNN-based branches are employed to extract local features followed by the transformer encoder blocks to capture long-range dependencies in the image (Wang et al. 2022a). In another technique, Wang and coauthors used a CNN and a transformer branch to learn local and global image representations, which were then integrated to generate the final output. Their approach demonstrated significant improvement as compared to other existing approaches. Hampali and co-authors developed a hybrid pose estimation method, named Keypoint Transformer (Hampali et al.

2021). In the proposed method they utilized both CNN and transformer-based modules to efficiently estimate human joints as 2D keypoints. Experimental results showed exemplary results of this approach on datasets including InterHand2.6M.

# 5. Challenges

ViTs and HVTs have demonstrated exceptional performance not only in computer vision, but also in various other domains. Nonetheless, integrating convolutional operations effectively into the transformer architecture poses several challenges for HVTs. Some of the ViT and HVT challenges include:

- Humans have a capacity for leveraging rotation invariance and equivariance to comprehend the world within its context. However, deep learning models encounter difficulties in upholding equivariance/invariance as required. The challenge to equip ViTs with the appropriate inductive biases to effectively represent objects and video frames in an equivariant fashion is undeniably complex.

- The MSA mechanism in ViTs and the convolution operation in CNNs both rely on dense matrix multiplication to capture data dependencies. Therefore, both ViT and HVT architectures (CNN-Transformers) may face high computational complexity and memory overhead. As a result, they may encounter challenges when attempting to model dense applications such as volumetric analysis and segmentation.

- Training ViTs and HVTs generally require powerful hardware resources like GPUs due to their computational complexity. This can limit their deployment in real-world applications, especially on edge devices, due to the hardware constraints and associated costs.

- The intricate architecture of ViT poses challenges for its interpretability, particularly in the context of multi-head attention, where attention weights from each layer are intricately blended. As a result, comprehending which segments of the input tokens are actively involved in the decision-making process proves to be challenging.
- A major challenge faced by HVT architectures is the efficient merging of learned features from both transformer and convolutional layers. While the transformer layers learn global features that are independent of spatial location, convolutional layers learn local features that are spatially correlated. In architectural terms, the efficient unification of MSA and CNN layers can potentially result in improved performance in various vision tasks.
- ViTs and HVTs can process complex image data accurately due to their high learning capacity. However, this also means that they require large training datasets to effectively learn and generalize from the data. This poses a challenge, particularly in the medical image domain, where obtaining a large amount of annotated data is often difficult and time-consuming. The need for obtaining extensive labeled data can be a significant obstacle, consuming valuable resources and time, and impeding the development and application of HVTs in medical imaging.

# 6. Future directions

Understanding how ViTs make decisions is crucial due to their significant capabilities. As ViTs evolve, comprehending their inner workings becomes even more important. Knowing how the model functions is essential for understanding its decision-making instead of treating it as a black box. To improve ViTs, future research can explore different ways of interpreting the model. These methods include visualizing features, generating activation maps, and using techniques like Grad-CAM. Such approaches are vital for understanding the model's behavior and improving its transparency. By revealing the decision-making process, biases and errors can be identified and addressed more effectively. This involves focusing on specific regions rather than arbitrary ones, improving model development and deployment.

ViTs and HVTs are large models with billions of parameters, which necessitates the need for lightweight architectures. Their high complexity may lead to latency in inference and significant overhead on energy consumption. There is a need to explore new and innovative design principles for efficient ViTs and HVTs with significant inference rates to enable their practical deployment in real-world applications, edge devices, and computationally limited systems, such as satellites. Knowledge distillation emerges as a promising approach to generate data-efficient and compact models by transferring knowledge from high-capacity models to simpler ones.

HVTs combine the strengths of CNNs and transformers, making significant advancements in image analysis and computer vision. However, to fully utilize their potential, it is important to explore suitable ways in which the convolution and self-attention mechanisms can be integrated for specific vision applications. This involves an in-depth analysis of integration methods based on their suitability for various contexts, such as early layer integration, lateral layer integration, sequential integration, parallel integration, hierarchical integration, attention-based integration, and attention-based integration.

The HVT's local and global processing capabilities make them quite promising for a wide range of vision applications, with potential benefits beyond vision-related tasks. To further enhance the performance of HVTs, it is important to gain a deeper understanding of image content and associated operations, which can help in devising better hybrid and deep architectures. The investigation of the potential utilization of hand-crafted operators in combination with the hybrid and dynamic feature extraction mechanisms of CNN-Transformer architectures may be particularly important in the near future. Developing new and effective blocks using both convolution and self-attention mechanisms is also a promising area for research.

In summary, the future of HVTs looks bright, with immense potential for various applications in the field of image analysis, computer vision, etc. In our opinion, it is better to also focus on possible integration methods that merge self-attention and convolution layers within HVT architectures for specific vision tasks. This focus should also extend to understanding image content and operations, developing effective blocks that combine convolution and self-attention, and utilizing multimodality and multitasking in ViT and HVT architectures.

## 7. Conclusion

The ViT has gained considerable attention in research due to its promising performance in specific image-related tasks. This success can be attributed to the MSA module integrated into ViT architectures, enabling the modeling of global interactions within images. To enhance their performance, various architectural improvements have been introduced. These improvements can be categorized as patch-based, knowledge distillation-based, attention-based, multi-transformer-based, and hybrid approaches. This paper not only examines the architectural taxonomy of ViTs but also explores the fundamental concepts underlying ViT architectures.

While ViTs have impressive learning capacities, they may suffer from limited generalization in some applications due to their lack of inductive bias that can capture local relations in images. To address this, researchers have developed HVTs, also known as CNN-Transformers, which leverage both self-attention and convolution mechanisms to learn both local and global information.

Several studies have proposed ways to integrate convolution-specific inductive bias into transformers to improve their generalization and capacity. Integration methodologies include early-layer integration, lateral-layer integration, sequential integration, parallel integration, hierarchical integration, and channel boosting-based integration. In addition to introducing taxonomy for HVT architectures based on their integration methodology, we also provide an overview of how they are used in various real-world computer vision applications. Despite current challenges, we believe that HVTs have enormous potential due to their capability to perform learning at both local and global levels.

## Acknowledgments

This work has been conducted at the pattern recognition lab, Pakistan Institute of Engineering and Applied Sciences, Islamabad, Pakistan. We extend our sincere gratitude to Dr. Abdul Majid and Dr. Naeem Akhter of DCIS, PIEAS for their invaluable assistance in improving the manuscript. Additionally, we acknowledge Pakistan Institute of Engineering and Applied Sciences (PIEAS) for a healthy research environment which led to the work presented in this article.

## Competing interests

The authors declare no competing financial and/or non-financial interests about the described work.